\documentclass[11pt]{article}
\usepackage[final]{acl}

\usepackage{times}
\usepackage{latexsym}
\usepackage[T1]{fontenc}
\usepackage[utf8]{inputenc}
\usepackage{microtype}
\usepackage{inconsolata}
\usepackage{graphicx}
\usepackage{booktabs}
\usepackage{array}
\usepackage{amssymb}
\usepackage{amsmath}
\usepackage[most]{tcolorbox}
\usepackage{placeins}
\usepackage{xcolor}
\usepackage{xurl}
\usepackage{hyperref}

\definecolor{tickgreen}{rgb}{0.0,0.5,0.0}
\newcommand{\yes}{\textcolor{tickgreen}{\checkmark}}
\newcommand{\no}{\textcolor{black!30}{\textendash}}

\newtcolorbox{promptbox}[1][]{
 colback=gray!5, colframe=gray!50, boxrule=0.5pt, sharp corners,
 fontupper=\small\ttfamily, title=#1, fonttitle=\bfseries\small\sffamily,
 coltitle=black, colbacktitle=gray!20, breakable,
 left=4pt, right=4pt, top=4pt, bottom=4pt
}

\newtcolorbox{promptboxnb}[1][]{
 colback=gray!5, colframe=gray!50, boxrule=0.5pt, sharp corners,
 fontupper=\small\ttfamily, title=#1, fonttitle=\bfseries\small\sffamily,
 coltitle=black, colbacktitle=gray!20, breakable=false,
 left=4pt, right=4pt, top=4pt, bottom=4pt
}

\title{\textsc{DiagFlowBench}: Evaluating How Language Models Handle\\ Off-Procedure Inputs in Grounded Diagnostic Dialogue}

\author{
 \textbf{Guillermo Gil de Avalle\textsuperscript{1}}, \quad
 \textbf{Laura Maruster\textsuperscript{1}}, \quad
 \textbf{Shaina Raza\textsuperscript{2}}, \quad
  \textbf{Christos Emmanouilidis\textsuperscript{1}} \\
  \textsuperscript{1} University of Groningen, PO Box 72, 9700 AB Groningen, Netherlands \\
  \textsuperscript{2} Vector Institute for Artificial Intelligence, W1140-108 College Street, Toronto, Canada \\
 \small{\texttt{\{g.gil.de.avalle,l.maruster,c.emmanouilidis\}@rug.nl}, \texttt{shaina.raza@vectorinstitute.ai}}
}

\begin{document}
\maketitle

\begin{abstract}
Language models increasingly serve as advisory systems in maintenance operations. To prevent hallucination, established approaches ground these models in procedural documentation, constraining them to prescribed sequences. In practice, however, operators may stray from these steps, requiring models to recognise unscripted, off-procedure utterances. Current benchmarks rarely prioritise this capability. We introduce \textsc{DiagFlowBench}\footnote{Code and data are released at \url{https://github.com/guille-gil/DiagFlowBench}.}, a dataset of 50 industrial diagnostic flowcharts from a consumer manufacturer converted into 1{,}676 multi-turn conversations that contrast compliant with off-procedure utterances. Evaluating a panel of ten commercial and open-weight models reveals high variability in abstention rates, with models often selecting a real but contextually inadequate step rather than fabricating facts. The inherent plausibility and authority of this mapped but wrong advice exposes a challenging vulnerability for grounding systems.

\end{abstract}

\section{Introduction}

Large language models (LLMs) are increasingly deployed as maintenance advisory systems \citep{kernanfreire2023cognitive,colabianchi2024assistant}, helping operators through natural language to consult procedural documentation to propose the next action based on provided observations \citep{turner2019decision,deng2024prescription}. Industrial diagnostic procedural knowledge is typically captured as flowcharts or decision trees \citep{vidyaratne2024troubleshooting}, structures that recent work uses to ground maintenance systems \citep{wang2024ontology,emmanouilidis2019human}. Because such flowcharts enumerate admissible sequences of steps, mapping LLM outputs to this closed set acts as a built-in safeguard against hallucination, ensuring traceability and faithfulness to the procedure \citep{poesia2022synchromesh,maynez2020faithfulness}.

The challenge, however, is that operators rarely request advice in the exact terms of the procedure \citep{antonovsky2014maintenance}. As a conversation unfolds, operators may report novel symptoms or raise questions mid-inspection that the documented steps do not anticipate. This difficulty is compounded by the natural language operators produce, which typically features dense text with abbreviations, shorthand, and domain-specific terminology that resists clean mapping onto documentation \citep{brundage2021tlp,dima2021adapting}. Experience widens the gap, as seasoned operators draw on tacit knowledge of the equipment rather than the literal manual \citep{dreyfus2004skill}, making their queries further diverge from the mapped steps. 

In practice, an LLM-based advisory system may receive these off-procedure queries mid-dialogue, mixed among on-procedure turns, and must handle both during generation. The two situations have tended to be evaluated independently. Flowchart-grounded benchmarks typically evaluate compliance on clean, mappable inputs \citep{raghu2021flodial,zhang2025pfdial}, while off-procedure recognition is typically evaluated as abstention over single-turn queries \citep{feng2024abstain,larson2019clinc150}. A live diagnostic dialogue presents both at once, demanding that a model advance through the procedure on compliant turns while recognising, in the same conversation, the turns that leave it.

To close this gap, we introduce \textsc{DiagFlowBench}\footnote{This work has been accepted at the Conference on Empirical Methods in Natural Language Processing (EMNLP 2026), Industry Track.} to evaluate flowchart-grounded models under both cooperative and off-procedure conditions. Our contributions are (1) a dataset of 50 anonymised industrial flowcharts from a consumer manufacturer rendered into 1{,}676 multi-turn conversations pairing clean dialogues with off-procedure counterparts, (2) an evaluation of ten diverse commercial and open-weight models, and (3) the identification of a failure mode, which we term \textit{forced mapping}, whereby models forcefully assign an off-procedure query to an inapplicable step. Because these returned steps are genuine procedure nodes, similarity-based grounding alone struggles to flag them as failures. Ultimately, our findings reveal that the constraint mechanisms designed to prevent hallucination may instead camouflage contextual errors as valid steps, turning a built-in safety feature into a blind spot.

\section{Related Work}
\label{sec:related}

Task-oriented dialogue (TOD) systems guide a user toward a goal across multiple turns, traditionally by tracking slots and intents over a fixed schema \citep{budzianowski2018multiwoz}. Flowchart-grounded dialogue specialises this setting by replacing the schema with an explicit procedure graph, so that each turn advances along a documented troubleshooting workflow \citep{raghu2021flodial,zhan2023flowchart}. Benchmarks in this line primarily measure how faithfully a model follows that workflow. FloDial established this task by rendering troubleshooting flowcharts into conversations, scoring models on their ability to select the specific steps licensed by operator queries \citep{raghu2021flodial}. Later work pursued scale, deriving procedures from UML diagrams \citep{zhang2025pfdial} and generating synthetic dialogues from existing troubleshooting flowcharts to relieve data scarcity \citep{zhan2023flowchart}. More recently, GuideBench tests compliance for specific domain rules \citep{diao2025guidebench}, while SOP-Bench evaluates whether a model is able to adhere to industrial operating procedures and execute discrete tool calls \citep{nandi2025sopbench}. Despite these variations, this body of work largely assumes that every operator query maps cleanly to the documented steps, and often fails to account for utterances that fall outside its scope.

This underlying recognition skill is studied independently as abstention. Traditional abstention detection frames this challenge around single inputs presented in isolation \citep{larson2019clinc150,rajpurkar2018squad}. However, even in these isolated settings, abstention remains a persistent challenge for LLMs, as they routinely misjudge the limits of their own knowledge \citep{yin2023know} and exhibit a systemic bias toward generating an answer rather than declining \citep{feng2024abstain}. This tendency persists under retrieval paradigms, where models synthesise answers from passages that offer no actual support for the query \citep{cuconasu2024noise}, thereby undermining retrieval as a safeguard against hallucination \citep{shuster2021retrieval}. When this challenge is extended to multi-turn TOD dialogues, off-procedure inputs are primarily treated as interference. For example, injecting casual conversation into MultiWOZ slot-filling dialogues actively degrades state tracking and task success \citep{budzianowski2018multiwoz,stricker2024chitchat}. Other benchmarks frame these departures as explicit rejection tasks. CGoDial inserts out-of-scope turns that the system must detect and meet with a fixed default reply \citep{dai2022cgodial}. Meanwhile, FlowAgent scores workflow agents strictly on their ability to remain compliant with a procedure when a user request strays \citep{shi2025flowagent}. In all these setups, the off-procedure turn is treated purely as a binary classification task to be set aside or ignored. Determining what a grounded model returns when it fails to abstain, rather than merely whether it abstained, requires evaluating each response against a specific procedure position. Without both the procedure and the position, a genuine but inapplicable step is indistinguishable from a correct one. 

\begin{table}[t]
\centering
\small
\setlength{\tabcolsep}{4pt}
\renewcommand{\arraystretch}{1.15}
\begin{tabular}{lccccc}
\toprule
 & Graph & Obs. & Docs & Multi & Off \\
\midrule
FloDial               & \yes & \yes & \yes & \yes & \no  \\
PFDial                & \yes & \no  & \no  & \yes & \no  \\
GuideBench            & \no  & \no  & \yes & \no  & \no  \\
CGoDial               & \yes & \no  & \no  & \yes & \yes \\
FlowAgent             & \yes & \no  & \no  & \yes & \yes \\
SOP-Bench             & \yes & \no  & \yes & \no  & \no  \\
\midrule
\textbf{\textsc{DiagFlowBench}}& \yes & \yes & \yes & \yes & \yes \\
\bottomrule
\end{tabular}
\caption{Procedural and flowchart-grounded benchmarks against the five properties an off-procedure evaluation requires, defined in the text: graph-structured procedure (Graph), observation-driven turns (Obs.), realistic documentation (Docs), multi-turn dialogue (Multi),
and off-procedure inputs in evaluation (Off).}
\label{tab:benchmarks}
\end{table}

\begin{table*}[!t]
\centering
\small
\setlength{\tabcolsep}{5pt}
\renewcommand{\arraystretch}{1.2}
\begin{tabular}{>{\raggedright\arraybackslash}p{2.8cm}>{\raggedright\arraybackslash}p{3.4cm}>{\raggedright\arraybackslash}p{5.2cm}>{\raggedright\arraybackslash}p{3.4cm}}
\toprule
\textbf{Outcome} & \textbf{Condition} & \textbf{Description} & \textbf{Distinguished by prior work} \\
\midrule
Fabrication (FA) & $\hat{a}_t \notin V$ &
The model names a step that does not exist in the procedure &
Yes, as extrinsic hallucination \citep{maynez2020faithfulness,ji2023survey} \\
Forced mapping (FM) & $\hat{a}_t \in V$, no edge into $\hat{a}_t$ entailed by $o_t$ &
The model names a real procedure step that the operator input gave no reason to select &
No \\
Correct abstention (CA) & no $\hat{a}_t$ is named &
The model recognises the operator input falls outside the procedure and redirects, escalates, or asks for clarification &
Partially, as single-turn abstention \citep{larson2019clinc150,feng2024abstain} \\
\bottomrule
\end{tabular}
\caption{The three off-procedure outcomes and how prior evaluation treats each. Conditions are stated for turns where no outgoing edge of $v_t$ is entailed by $o_t$.}
\label{tab:failure_modes}
\end{table*}

An off-procedure report to a grounded advisor sits precisely where these two research lines intersect. To accurately read how a model handles such queries, we specify five evaluation conditions. The procedure must operate as a \textit{Graph}, so that inputs can fall outside structural edges. Turns must be \textit{Observation-driven} to force inference over direct reading. The documentation should ideally be \textit{Realistic}, or at least driven from real documentation, to faithfully represent the complexity of real-world deployments. The dialogue must be \textit{Multi-turn} to ensure active position tracking difficulty. Because models rarely self-correct after committing to an erroneous path \citep{laban2025lost}, a multi-turn setup is necessary to capture how the degradation from a single mishandled off-procedure turn cascades through subsequent interactions. Finally, \textit{Off-procedure} inputs must be explicitly present to create turns lacking any mappable response. Table~\ref{tab:benchmarks} assesses the aforementioned benchmarks against these five criteria, alongside \textsc{DiagFlowBench}.

\section{Task Formulation}
\label{sec:task}
We model a procedure as a directed graph $G = (V, E)$ whose nodes are diagnostic steps and whose edges are admissible transitions. Each edge $e = (v \rightarrow v')$ carries an observation $\ell(e)$ that licenses the move to $v'$, so the admissible inputs at a node $v$ are the labels on its outgoing edges, $L(v) = \{\ell(e) : e = (v \rightarrow \cdot)\}$. Some nodes are terminal, i.e. they mark the end of the procedure.

A conversation visits nodes in sequence. At the current node $v_t$ the operator produces an utterance $o_t$, and the model, given the history $h_t$, must name the next action $\hat{a}_t$. The utterance is on-procedure when it entails the label on an outgoing edge of $v_t$, that is $o_t \models \ell(e)$ for some $e = (v_t \rightarrow \cdot)$, and the target of that edge is then the correct next action. It is off-procedure when it entails none,
\[
o_t \not\models \ell(e)\quad\text{for all } e=(v_t\rightarrow\cdot),
\]
in which case no node of $G$ is a correct answer. Treating this as no valid response, rather than a missing one, is the closed-world reading of the procedure \citep{reiter1978cwa}. Operator turns follow a fixed reference path rather than reacting to $\hat{a}_t$, so every model sees the same conversation and on- and off-procedure turns can be scored independently \citep{raghu2021flodial}. Appendix~\ref{app:notation} collects the interpretation of notation.

\paragraph{On-procedure capability.} During an on-procedure turn the model must locate $v_t$ from $h_t$, choose the successor that $o_t$ entails, and end at a terminal node. These define two capabilities measured per turn in the evaluation protocol. Write $v^{*}_t$ for the target of the edge out of $v_t$ whose label $o_t$ entails, and $\rho(\hat{a}_t)$ for the node a response resolves to under the reference-resolution step of Appendix~\ref{app:judge}, which returns $\varnothing$ when the response names no node. Step accuracy (SA) is whether the named $\hat{a}_t$ is the successor that $o_t$ entails,
\[
\mathrm{SA}(t)\;=\;\mathbf{1}\!\left[\rho(\hat{a}_t)=v^{*}_t\right].
\]
Termination recognition (TR) is whether the model halts at a terminal node, evaluated at terminal $v_t$ alone,
\[
\mathrm{TR}(t)\;=\;\mathbf{1}\!\left[\rho(\hat{a}_t)=\varnothing\right].
\]
On-procedure behaviour is therefore the baseline against which off-procedure behaviour is read.

\paragraph{Off-procedure failure modes.} During an off-procedure turn no outgoing edge of $v_t$ is entailed. The model should recognise that nothing in $L(v_t)$ is entailed by $o_t$ and decline to ground an action, by redirecting, escalating, or seeking clarification. This response is defined as correct abstention (CA). Every other response falls into one of two failure categories. In fabrication (FA), the model returns $\hat{a}_t \notin V$, a step absent from the procedure. In forced mapping (FM), the model returns a genuine node $\hat{a}_t \in V$, yet no edge into $\hat{a}_t$ is entailed by $o_t$. The step exists in the procedure, but the operator utterance simply gave no reason to select it. Table~\ref{tab:failure_modes} sets the three outcomes against the treatment each has received in prior work.
Separating FA from FM is what reveals the dissociation we report, namely that FA is rare and stable across the panel while FM is common and variable. FM falls outside common hallucination taxonomies \citep{maynez2020faithfulness,ji2023survey}. An FM selection is a genuine, graph-consistent step applied to the wrong context. This consistency makes FM uniquely difficult to detect. Standard grounding checks confirm the node is structurally valid but fail to verify its contextual relevance.

\section{The \textsc{DiagFlowBench} Benchmark}
\label{sec:construction}

\begin{figure}[tb]
 \centering
 \includegraphics[width=\columnwidth]{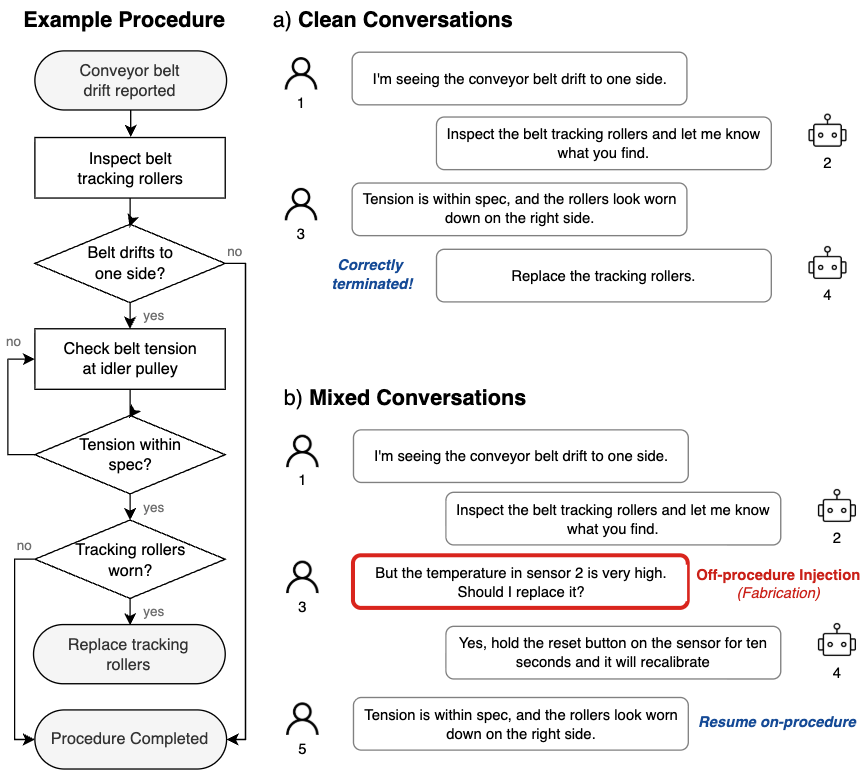}
 \caption{Task structure. A cooperative conversation (b) reports observations that map onto outgoing edges of the current node in the procedure (a). A mixed conversation (c) injects an off-procedure turn at a random position and then resumes on the original path.}
 \label{fig:task_structure}
\end{figure}

\textsc{DiagFlowBench} comprises 50 anonymised industrial diagnostic flowcharts rendered into 1{,}676 multi-turn conversations, all in English. Half are cooperative and half carry off-procedure injections. The two are built in pairs, so a mixed conversation differs from its cooperative counterpart only in whether the operator leaves the procedure, as observed in Figure~\ref{fig:task_structure}.

\subsection{Procedures}
The source procedures consist of troubleshooting and operator-control flowcharts extracted from the maintenance documentation of a consumer manufacturer. These cover diverse fault classes, including conveyor belt tracking, vision inspection calibration, robotic placement, and CNC spindle acceptance. 

To safely share this data, we anonymised the flowcharts using a three-step pipeline that removes all company identifiers and rewrites each procedure into a neutral industrial domain. This process strictly preserves the original diagnostic logic and graph topology. Subject-matter experts at the partner organisation confirmed that no proprietary identifiers survived the substitution. Furthermore, they verified that all branching logic, temporal dependencies, and escalation paths remain unchanged. Following these successful validation checks, the organisation approved the open release of the flowcharts. The resulting graphs contain between 7 and 60 nodes, a mean of 31, and decision branching factors ranging from 2 to 4 options. Appendix~\ref{app:dataset} illustrates the complete anonymisation pipeline and provides comprehensive dataset statistics.

\subsection{Conversations}
For each graph we enumerate root-to-terminal paths under a greedy set cover that covers every decision branch, capping paths per graph to avoid over-sampling large graphs. Appendix~\ref{app:dataset} provides the path-selection procedure. Each path is rendered into an operator script by prompting a model with the current node and target edge label, after which a verification pass repairs any utterance that leaks downstream steps or violates edge semantics, and human annotators review the output. A noise-perturbation pass produces a second variant that preserves the process topology and path while introducing the typos, telegraphic phrasing, and abbreviated terms characteristic of maintenance work-order text \citep{brundage2021tlp}, following the robustness-testing protocol of \citet{liu2021robustness}. Appendix~\ref{app:gen_prompts} depicts the generation and validation prompts, as well as the pipeline details.

Thereafter, we build the mixed counterpart of each conversation by injecting off-procedure turns at non-terminal positions, each generated with the same model and screened by a zero-temperature verifier. However, off-procedure inputs are not homogeneous. Previous answerability work shows that distinguishing why an input cannot be answered, rather than treating all such inputs alike, matters for how a system should respond \citep{ravichander2019privacyqa}. We therefore propose stratifying injections into three categories along their relationship to the procedure: \textit{Coverage Gap} (a symptom the procedure has no branch for), \textit{Undocumented Malfunction} (a fault outside any branch condition at the current node), or \textit{Unrelated Question} (a query outside the diagnostic task entirely). Appendix~\ref{app:gen_prompts} offers details on the injection process and prompts, and Appendix~\ref{app:scoring} provides details and examples of the stratification.

\section{Evaluation Protocol}
\label{sec:eval}

\subsection{Models}
\label{sec:models}
 We evaluate ten models that span commercial and open-weight releases. The commercial set is Gemini~2.5~Flash \citep{geminiteam2025gemini} and GPT-4o~Mini \citep{openai2024gpt4omini}. The open-weight set comprises three Llama models spanning two generations, Llama~4~Scout and Llama~4~Maverick \citep{meta2025llama4} alongside Llama~3.3~70B \citep{grattafiori2024llama}, together with Qwen3~235B and Qwen3~30B in thinking mode \citep{yang2025qwen3}, Mistral~Small~24B \citep{mistral2025small3}, GPT-OSS~120B \citep{openai2025gptoss}, and Nemotron~3~Super~120B \citep{nvidia2026nemotron}. The panel isolates three analytical dimensions. The access-tier dimension contrasts two commercial models with eight open-weight releases. The architecture dimension spans instruction-tuned or instruct, mixture-of-experts (MoE), and reasoning designs across providers. A scale variation within the Llama family completes the set. Pinned models, parameter counts, architecture, and hyperparameters used per model are specified in Appendix~\ref{app:eval_prompts}. 
 
\subsection{Prompting}
\label{sec:prompting}
The system prompt casts the model as a maintenance troubleshooting assistant, supplies the procedure graph as structured JSON, and requests one next action per turn with an explicit stop at a terminal node. The prompt also instructs the model to abstain when the reported observation matches no node. However, this instruction is one among several, mirroring a deployment reality in which a system prompt serves multiple objectives at once. Appendix~\ref{app:counterfactual} reports a choice-framed abstention ablation in which abstention detection is made the only task objective, isolating the share of FM that survives that change. Moreover, the procedure graph is supplied directly in the prompt to isolate the grounding capabilities of the model. All calls use temperature zero. The per-turn output limits (which differ for reasoning models) and the evaluation prompt are detailed in Appendix~\ref{app:eval_prompts}.

\subsection{Per-turn scoring}
\label{sec:scoring}
The fixed reference path lets on- and off-procedure turns be scored independently. On-procedure turns are matched to a node by word-level Jaccard similarity at a threshold calibrated on a held-out 90/10 graph split, from which SA and TR follow directly (Appendix~\ref{app:judge}).

Off-procedure responses are free-form prose that neither matches a node cleanly nor carries an unambiguous abstention marker. Therefore, we classify them individually into CA, FM, or FA using a fixed language-model judge \citep{zheng2023judge}. We use the \texttt{claude-haiku-4-5} model \citep{anthropic2025haiku45}, whose provider is absent from the evaluated panel to avoid self-preference bias \citep{panickssery2024self}. Although conversations and injections were also generated with Anthropic models, the judge performs its task against the outputs of the evaluated models alone, not the conversations, preventing overlap that drives the bias. We also asked two human annotators to validate a stratified sample of classifications, yielding a human-model agreement of $\kappa = 0.79$ alongside a human-human agreement of $\kappa = 0.83$ \citep{landis1977measurement}. We provide further details about the judge in Appendix~\ref{app:judge}. Since the disagreements that remain concentrate on the FM-versus-CA boundary, Appendix~\ref{app:band} propagates them to the reported splits, bounding how far the FM and CA of each model could shift in the worst case.

\section{Results and Discussion}
\label{sec:results}

\subsection{On-procedure Competence}
As observed in Table~\ref{tab:main_results}, SA ranges from 70.1 to 85.0\% across the panel. This provides evidence of relatively high performance when the operator query is covered by the manual. TR appears to be the much weaker and more volatile capability, spanning from 38.8 to 98.5\%. Consequently, a model capable of selecting the correct next step may still fail to accurately predict procedural termination. 

\begin{table}[t]
\centering
\small
\setlength{\tabcolsep}{3.5pt}
\caption{Results per model (all values in \%). \textbf{Bold}: best per column (highest for SA, TR, CA; lowest for FA, FM). SA\,=\,step accuracy; TR\,=\,termination recognition; FA\,=\,fabrication; FM\,=\,forced mapping; CA\,=\,correct abstention.}
\label{tab:main_results}
\begin{tabular}{lcccrrr}
\toprule
\textbf{Model} &
  \multicolumn{2}{c}{\textbf{On-proc.}} & &
  \multicolumn{3}{c}{\textbf{Off-proc.}} \\
\cmidrule{2-3} \cmidrule{5-7}
 & SA & TR & & FA & FM & CA \\
\midrule
\multicolumn{7}{l}{\emph{Commercial}}\\
Gemini 2.5 Flash  & 84.0 & 71.8 & & 2.4 & 25.6 & 72.0 \\
GPT-4o Mini       & 75.4 & 38.8 & & 5.3 & 36.4 & 58.3 \\
\midrule
\multicolumn{7}{l}{\emph{Open-weight}}\\
Qwen3 235B        & 73.7 & \textbf{98.5} & & 6.3 & 28.8 & 64.9 \\
Qwen3 30B         & 70.1 & 70.1 & & 5.1 & 67.4 & 27.4 \\
Mistral Small 24B & 73.6 & 79.6 & & 5.3 & 27.6 & 67.1 \\
GPT-OSS 120B      & 76.2 & 74.4 & & \textbf{2.2} & 43.1 & 54.7 \\
Nemotron 3 Super  & 70.3 & 86.1 & & 4.1 & 22.1 & 73.9 \\
\midrule
\multicolumn{7}{l}{\emph{Scalability Test (Llama family)}}\\
Llama 4 Scout     & 79.2 & 61.6 & & 8.6 & 51.9 & 39.4 \\
Llama 4 Maverick  & 82.8 & 72.8 & & 5.0 & 29.3 & 65.7 \\
Llama 3.3 70B     & \textbf{85.0} & 83.4 & & 3.0 & \textbf{15.7} & \textbf{81.3} \\
\bottomrule
\end{tabular}
\end{table}

\subsection{Off-procedure Competence}
\begin{figure*}[t]
 \centering
 \includegraphics[width=\linewidth]{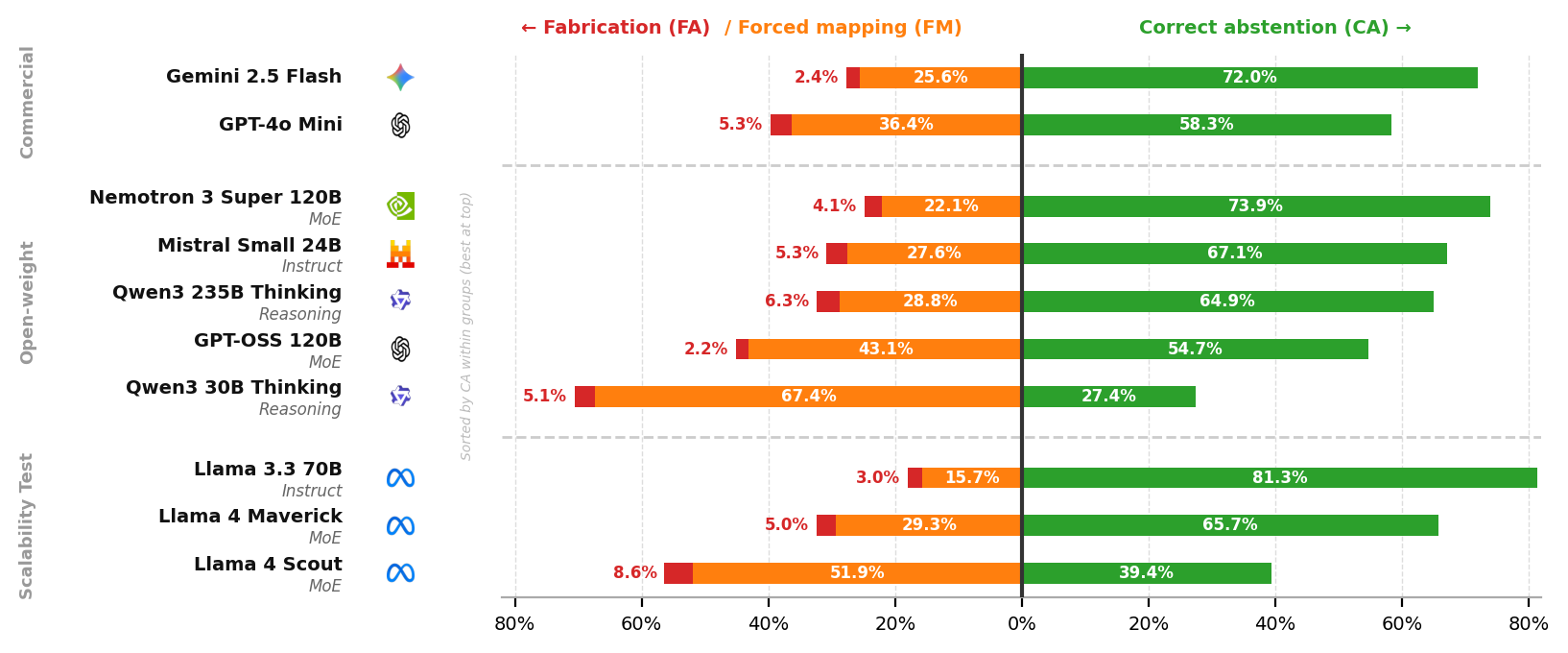}
 \caption{Off-procedure outcomes per model, sorted by correct abstention. Correct abstention (green, right) is the target; FM (orange, left) and FA (red, far left) are failures.}
 \label{fig:diverging}
\end{figure*}

Consistent with the on-procedure results, FA remains strictly bounded between 2.2 and 8.6\%, confirming that procedural grounding effectively prevents models from inventing steps. In sharp contrast, FM exhibits substantially higher and more volatile rates across the panel (15.7 to 67.4\%), emerging as the dominant failure mode. Notably, a low FA rate provides no guarantee of overall robustness. For instance, GPT-OSS~120B pairs the lowest FA score with an FM rate exceeding 40\%, demonstrating that a model can successfully minimise FA while remaining faulty.  Moreover, while CA remains the most frequent response for eight of the ten models, Figure~\ref{fig:diverging} illustrates that failure rates remain non-negligible, with even the highest-performing model failing to abstain on 18.7\% of off-procedure utterances. These results reveal an error profile in which FM operates as the primary hallucination mode while extrinsic FA remains rare.

Neither access tier nor parameter count reliably predicts CA within this panel. Commercial models land in the middle of the panel, whereas the open-weight group spans a 54-point variability in CA. Parameter counts appear disconnected from performance on this metric as well. For instance, the smaller Mistral Small~24B achieves a higher CA rate than the much larger Qwen3~235B. Additionally, the scalability test within the Llama family reveals that generational advances do not necessarily improve off-procedure handling, as the older Llama~3.3~70B outperforms both newer Llama~4 models. Finally, the model architecture presents the only tentative pattern. Instruction-tuned models (such as Llama~3.3~70B and Mistral Small~24B) perform relatively well despite not being the largest or newest. Nevertheless, they remain susceptible to FM, exhibiting error rates between 15.7 and 27.6\%. Furthermore, although reasoning-oriented architectures generate an explicit scratchpad prior to output to structure their reasoning, Qwen3~30B records the lowest CA rate in the entire panel, namely 27.4\%. Its much larger counterpart, Qwen3~235B, avoids this extreme but similarly fails to surpass standard instruction-tuned models.

Model scale, architecture, and parameter count therefore fail to definitively predict off-procedure reliability. Contrasted with the relatively high SA baseline, the off-procedure track exposes variable levels of model brittleness when facing unexpected operator inputs. Based on these findings, we argue that FM is fundamentally an \textit{over-compliance} failure. The underlying drive toward task completion actively overrides the capacity of the model to recognise when an input falls outside the procedural scope.

\subsection{Drivers and Outcomes of Forced Mapping}
\label{sec:per_cat}

We measure FM under a node-selection regime, in which naming a procedure node is the standing instruction (\S\ref{sec:prompting}). This introduces a tension between instruction compliance and abstention, as the primary directive to advance the workflow actively competes with the conditional safeguard to reject inapplicable inputs. Appendix~\ref{app:counterfactual} demonstrates that reframing the prompt as a task whose objective is solely detecting abstention reduces FM, but still leaves a substantial residual. The baseline prompt therefore elevates the measured error without solely producing the phenomenon. The per-category spread demonstrates that this error is not simply blind prompt compliance. Under identical instructions, models successfully abstain on up to 98.5\% of \textit{unrelated questions} but systematically force-map \textit{coverage-gap} inputs. This categorical ordering holds across all models and tracks how closely each
category of injection resembles the procedure text (Appendix~\ref{app:scoring}),
which points to lexical similarity between the off-procedure query and the
procedure as the driver of FM. 


Once an FM is triggered, the error is rarely confined to the turn that produces it. We define \textit{Recovery Rate} as the fraction of on-procedure turns immediately following a failure where the model successfully names the correct next step. Across the panel, this recovery rate plummets to between 1.0 and 9.1\%, a stark contrast to the 70.1 to 85.0\% SA baseline observed in clean conversations. A single mishandled injection therefore collapses the subsequent procedural tracking to near zero, consistent with findings in literature \citep{laban2025lost}. Because later turns carry richer context and score higher on clean conversations, turn position biases this measure upward (see Appendix~\ref{app:contamination}). The observed collapse therefore holds against that bias rather than resulting from it. Appendix~\ref{app:contamination} reports the recovery rates method and detailed results.

\section{Deployment Implications}
\label{sec:deployment}

FM is inherently harder to defend against than FA. Standard guardrails detect extrinsic hallucination by verifying source traceability. FM bypasses this verification entirely because the mapped steps are strictly genuine. This structural validity also misleads the operator. By confidently prescribing a documented step, FM may project an \textit{illusion of authority}, making the output appear plausible despite being contextually incorrect. As both automated checks and human intuition struggle to flag authentic steps in the wrong context, detection must actively validate the current utterance against the active node and its edges. Moreover, safety cannot be inferred from low FA rates either, as hallucination are not necessarily displaced but remapped.


Two properties shape FM mitigation in deployment. First, because FMs scale with lexical similarity to the manual, textual closeness acts as a risk indicator rather than a safety guarantee. This arguably penalises operators who naturally write using the procedural lexicon. Second, failure prevalence cannot be explained by model scale, architecture, or reasoning-oriented inference. Prompt design offers the only lever to reduce FM, but only when abstention becomes the sole task objective (Appendix~\ref{app:counterfactual}). However, even this format leaves a substantial residual, meaning the reduction in FM fails to offset the functional cost of removing other primary diagnostic instructions. Consequently, mitigation must shift from preventive constraints to active response validation. This validation cannot rely solely on generative models subject to the same lexical biases. Ultimately, while abstention prompting establishes a necessary baseline, safeguarding against FM requires external verification layers capable of strict contextual entailment to distinguish genuinely off-procedure inputs from superficial ambiguity.

\section{Conclusion}
\label{sec:conclusion}
We introduce \textsc{DiagFlowBench} to evaluate how maintenance advisory models handle off-procedure operator queries in flowchart-grounded diagnostic dialogue. Our findings reveal that constraining models to documented steps alone does not prevent failure, but instead swaps traditional hallucination for a failure mode where models select a real but contextually inadequate step. Ultimately, \textsc{DiagFlowBench} demonstrates that evaluating reliability in advisory systems requires measuring not just grounding compliance, but how safely a model fails when reality diverges from the manual.

\section*{Limitations}

This study is subject to several limitations. First, because real logs from a deployed advisory system were unavailable, the benchmark relies on synthetic conversations. Although generation and validation processes are in place to prevent it, synthetic language may still exhibit patterns absent from natural human speech. This could influence the measured effect, given the sensitivity of FM to lexical overlap. Furthermore, all evaluated procedures originate from a single industrial maintenance domain, restricting claims of cross-domain generality until an analogous evaluation is applied to other fields. The dataset is also exclusively in English. Because FM relies on surface-level word matching between utterance and procedure, applying this framework to languages with more complex morphological structures or distinct orthographies might reveal different mapping behaviours. Finally, every procedure is provided in full within the prompt. The results therefore speak to models that are given the procedure rather than ones that must retrieve it at runtime. While the released dataset can support future evaluations of inference-time retrieval setups, such configurations fall outside the scope of this paper.

\section*{Ethical Considerations}

The procedures originate in proprietary maintenance documentation and were anonymised before release through the pipeline of \S\ref{sec:construction}, with subject-matter experts confirming that no proprietary identifiers survive, so the released benchmark contains only the abstracted procedures. The benchmark is released under CC-BY-4.0, which permits commercial use, redistribution, and derivative work under attribution. Its intended use is the evaluation of procedural grounding, and the off-procedure conversations are a controlled stress test rather than a record of any implemented advisory system. The generation and evaluation prompts, code, and pinned model identifiers are released under MIT for reproducibility and auditing. The annotators who reviewed the generated conversations and validated the judge were doctoral researchers affiliated with the same institution as the research team and were compensated for their work.

\section*{Acknowledgments}
This research was funded by the European Union's Horizon Europe research and innovation programme under the AIXPERT project (Grant Agreement No.\ 101214389), which aims to develop an agentic, multi-layered, GenAI-powered framework for creating explainable, accountable, and transparent AI systems. 

\bibliography{custom}

@inproceedings{maynez2020faithfulness,
  title     = {On Faithfulness and Factuality in Abstractive Summarization},
  author    = {Maynez, Joshua and Narayan, Shashi and Bohnet, Bernd and McDonald, Ryan},
  booktitle = {Proceedings of the 58th Annual Meeting of the Association for Computational Linguistics (ACL)},
  pages     = {1906--1919}, year = {2020}, doi = {10.18653/v1/2020.acl-main.173}
}

@article{ji2023survey,
  title   = {Survey of Hallucination in Natural Language Generation},
  author  = {Ji, Ziwei and Lee, Nayeon and Frieske, Rita and Yu, Tiezheng and Su, Dan and Xu, Yan and Ishii, Etsuko and Bang, Yejin and Madotto, Andrea and Fung, Pascale},
  journal = {ACM Computing Surveys}, volume = {55}, number = {12}, pages = {1--38}, year = {2023}, doi = {10.1145/3571730}
}

@inproceedings{shuster2021retrieval,
  title     = {Retrieval Augmentation Reduces Hallucination in Conversation},
  author    = {Shuster, Kurt and Poff, Spencer and Chen, Moya and Kiela, Douwe and Weston, Jason},
  booktitle = {Findings of the Association for Computational Linguistics: EMNLP 2021}, pages = {3784--3803}, year = {2021}
}

@incollection{reiter1978cwa,
  title     = {On Closed World Data Bases},
  author    = {Reiter, Raymond},
  booktitle = {Logic and Data Bases}, editor = {Gallaire, Herv{\'e} and Minker, Jack}, pages = {55--76}, publisher = {Plenum Press}, address = {New York}, year = {1978}
}

@inproceedings{larson2019clinc150,
  title     = {An Evaluation Dataset for Intent Classification and Out-of-Scope Prediction},
  author    = {Larson, Stefan and Mahendran, Anish and Peper, Joseph J. and Clarke, Christopher and Lee, Andrew and Hill, Parker and Kummerfeld, Jonathan K. and Leach, Kevin and Laurenzano, Michael A. and Tang, Lingjia and Mars, Jason},
  booktitle = {Proceedings of the 2019 Conference on Empirical Methods in Natural Language Processing and the 9th International Joint Conference on Natural Language Processing (EMNLP-IJCNLP)},
  pages     = {1311--1316}, year = {2019}, doi = {10.18653/v1/D19-1131}
}

@inproceedings{rajpurkar2018squad,
  title     = {Know What You Don't Know: Unanswerable Questions for {SQuAD}},
  author    = {Rajpurkar, Pranav and Jia, Robin and Liang, Percy},
  booktitle = {Proceedings of the 56th Annual Meeting of the Association for Computational Linguistics (Volume 2: Short Papers)},
  pages     = {784--789}, year = {2018}, doi = {10.18653/v1/P18-2124}
}

@inproceedings{yin2023know,
  title     = {Do Large Language Models Know What They Don't Know?},
  author    = {Yin, Zhangyue and Sun, Qiushi and Guo, Qipeng and Wu, Jiawen and Qiu, Xipeng and Huang, Xuanjing},
  booktitle = {Findings of the Association for Computational Linguistics: ACL 2023}, pages = {8653--8665}, year = {2023}
}

@inproceedings{feng2024abstain,
  title     = {Don't Hallucinate, Abstain: Identifying {LLM} Knowledge Gaps via Multi-{LLM} Collaboration},
  author    = {Feng, Shangbin and Shi, Weijia and Wang, Yike and Ding, Wenxuan and Balachandran, Vidhisha and Tsvetkov, Yulia},
  booktitle = {Proceedings of the 62nd Annual Meeting of the Association for Computational Linguistics (Volume 1: Long Papers)},
  pages     = {14664--14690}, year = {2024}, doi = {10.18653/v1/2024.acl-long.786}
}

@inproceedings{cuconasu2024noise,
  title     = {The Power of Noise: Redefining Retrieval for {RAG} Systems},
  author    = {Cuconasu, Florin and Trappolini, Giovanni and Siciliano, Federico and Filice, Simone and Campagnano, Cesare and Maarek, Yoav and Tonellotto, Nicola and Silvestri, Fabrizio},
  booktitle = {Proceedings of the 47th International ACM SIGIR Conference on Research and Development in Information Retrieval}, year = {2024}, doi = {10.1145/3626772.3657834}
}

@inproceedings{budzianowski2018multiwoz,
  title     = {{MultiWOZ} -- A Large-Scale Multi-Domain {Wizard-of-Oz} Dataset for Task-Oriented Dialogue Modelling},
  author    = {Budzianowski, Pawe{\l} and Wen, Tsung-Hsien and Tseng, Bo-Hsiang and Casanueva, I{\~n}igo and Ultes, Stefan and Ramadan, Osman and Ga{\v{s}}i{\'c}, Milica},
  booktitle = {Proceedings of the 2018 Conference on Empirical Methods in Natural Language Processing (EMNLP)}, pages = {5016--5026}, year = {2018}
}

@inproceedings{raghu2021flodial,
  title     = {End-to-End Learning of Flowchart Grounded Task-Oriented Dialogs},
  author    = {Raghu, Dinesh and Agarwal, Shantanu and Joshi, Sachindra and {Mausam}},
  booktitle = {Proceedings of the 2021 Conference on Empirical Methods in Natural Language Processing (EMNLP)}, pages = {4348--4366}, year = {2021}, doi = {10.18653/v1/2021.emnlp-main.357}
}

@inproceedings{zhan2023flowchart,
  title     = {Turning Flowchart into Dialog: Augmenting Flowchart-grounded Troubleshooting Dialogs via Synthetic Data Generation},
  author    = {Zhan, Haolan and Maruf, Sameen and Qu, Lizhen and Wang, Yufei and Zukerman, Ingrid and Haffari, Gholamreza},
  booktitle = {Proceedings of the 21st Annual Workshop of the Australasian Language Technology Association (ALTA)}, year = {2023}
}

@inproceedings{zhang2025pfdial,
  title     = {{PFDial}: A Structured Dialogue Instruction Fine-tuning Method Based on {UML} Flowcharts},
  author    = {Zhang, Ming and Wang, Yuhui and Shen, Yujiong and Yang, Tingyi and Jiang, Changhao and Wu, Yilong and Dou, Shihan and Chen, Qinhao and Xi, Zhiheng and Zhang, Zhihao and Dong, Yi and Wang, Zhen and Fei, Zhihui and Wan, Mingyang and Liang, Tao and Ma, Guojun and Zhang, Qi and Gui, Tao and Huang, Xuanjing},
  booktitle = {Findings of the Association for Computational Linguistics: ACL 2025}, pages = {2626--2649}, year = {2025}
}

@inproceedings{diao2025guidebench,
  title     = {{GuideBench}: Benchmarking Domain-Oriented Guideline Following for {LLM} Agents},
  author    = {Diao, Lingxiao and Xu, Xinyue and Sun, Wanxuan and Yang, Cheng and Zhang, Zhuosheng},
  booktitle = {Proceedings of the 63rd Annual Meeting of the Association for Computational Linguistics (Volume 1: Long Papers)},
  pages     = {11361--11399}, year = {2025}, doi = {10.18653/v1/2025.acl-long.557}
}

@inproceedings{stricker2024chitchat,
  title     = {Chitchat as Interference: Adding User Backstories to Task-Oriented Dialogues},
  author    = {Stricker, Armand and Paroubek, Patrick},
  booktitle = {Proceedings of the 2024 Joint International Conference on Computational Linguistics, Language Resources and Evaluation (LREC-COLING)}, pages = {3203--3214}, year = {2024}
}

@inproceedings{dai2022cgodial,
  title     = {{CGoDial}: A Large-Scale Benchmark for {C}hinese Goal-oriented Dialog Evaluation},
  author    = {Dai, Yinpei and He, Wanwei and Li, Bowen and Wu, Yuchuan and Cao, Zheng and An, Zhongqi and Sun, Jian and Li, Yongbin},
  booktitle = {Proceedings of the 2022 Conference on Empirical Methods in Natural Language Processing (EMNLP)}, pages = {4097--4111}, year = {2022}, doi = {10.18653/v1/2022.emnlp-main.274}
}

@misc{shi2025flowagent,
  title = {{FlowAgent}: Achieving Compliance and Flexibility for Workflow Agents},
  author = {Shi, Yuchen and Cai, Siqi and Xu, Zihan and Qin, Yulei and Li, Gang and Shao, Hang and Chen, Jiawei and Yang, Deqing and Li, Ke and Sun, Xing},
  year = {2025}, eprint = {2502.14345}, archivePrefix = {arXiv}, primaryClass = {cs.CL}
}

@inproceedings{nandi2025sopbench,
  title     = {{SOP-Bench}: Complex Industrial {SOPs} for Evaluating {LLM} Agents},
  author    = {Nandi, Subhrangshu and Datta, Arghya and Nama, Rohith and Patel, Udita and Vichare, Nikhil and Bhattacharya, Indranil and Asija, Shivam and Gupta, Arushi and Carenini, Giuseppe and Xu, Jing and Ray, Shayan and Raja, Huzefa and Chan, Aaron and Carbone, Francesco and Fei, Esther Xu and Du, Gaoyuan and Akhtar, Zuhaib and Grover, Prince and Bhaduri, Sreyoshi and Chen, Weian and Zhang, Wei and Xiong, Ming},
  booktitle = {Proceedings of the ACM SIGKDD Conference on Knowledge Discovery and Data Mining (KDD)},
  year      = {2026}
}

@inproceedings{zheng2023judge,
  title     = {Judging {LLM-as-a-Judge} with {MT-Bench} and Chatbot Arena},
  author    = {Zheng, Lianmin and Chiang, Wei-Lin and Sheng, Ying and Zhuang, Siyuan and Wu, Zhanghao and Zhuang, Yonghao and Lin, Zi and Li, Zhuohan and Li, Dacheng and Xing, Eric P. and Zhang, Hao and Gonzalez, Joseph E. and Stoica, Ion},
  booktitle = {Advances in Neural Information Processing Systems (NeurIPS), Datasets and Benchmarks Track}, year = {2023}
}

@inproceedings{liu2023geval,
  title     = {{G-Eval}: {NLG} Evaluation using {GPT-4} with Better Human Alignment},
  author    = {Liu, Yang and Iter, Dan and Xu, Yichong and Wang, Shuohang and Xu, Ruochen and Zhu, Chenguang},
  booktitle = {Proceedings of the 2023 Conference on Empirical Methods in Natural Language Processing (EMNLP)}, pages = {2511--2522}, year = {2023}
}

@inproceedings{liu2021robustness,
  title     = {Robustness Testing of Language Understanding in Task-Oriented Dialog},
  author    = {Liu, Jiexi and Takanobu, Ryuichi and Wen, Jiaxin and Wan, Dazhen and Li, Hongguang and Nie, Weiran and Li, Cheng and Peng, Wei and Huang, Minlie},
  booktitle = {Proceedings of the 59th Annual Meeting of the Association for Computational Linguistics and the 11th International Joint Conference on Natural Language Processing (Volume 1: Long Papers)}, pages = {2467--2480}, year = {2021}
}

@article{landis1977measurement,
  title   = {The Measurement of Observer Agreement for Categorical Data},
  author  = {Landis, J. Richard and Koch, Gary G.},
  journal = {Biometrics}, volume = {33}, number = {1}, pages = {159--174}, year = {1977}
}

@article{antonovsky2014maintenance,
  title   = {Identification of the Human Factors Contributing to Maintenance Failures in a Petroleum Operation},
  author  = {Antonovsky, Ari and Pollock, Clare and Straker, Leon},
  journal = {Human Factors}, volume = {56}, number = {2}, pages = {306--321}, year = {2014}, doi = {10.1177/0018720813491424}
}

@article{dreyfus2004skill,
  title   = {The Five-Stage Model of Adult Skill Acquisition},
  author  = {Dreyfus, Stuart E.},
  journal = {Bulletin of Science, Technology \& Society}, volume = {24}, number = {3}, pages = {177--181}, year = {2004}, doi = {10.1177/0270467604264992}
}

@inproceedings{kernanfreire2023cognitive,
  title     = {Harnessing Large Language Models for Cognitive Assistants in Factories},
  author    = {Kernan Freire, Samuel and Foosherian, Mina and Wang, Chaofan and Niforatos, Evangelos},
  booktitle = {Proceedings of the 5th International Conference on Conversational User Interfaces (CUI)}, year = {2023}, doi = {10.1145/3571884.3604313}
}

@article{colabianchi2024assistant,
  title   = {Assessment of a large language model based digital intelligent assistant in assembly manufacturing},
  author  = {Colabianchi, Silvia and Costantino, Francesco and Sabetta, Nicol{\`o}},
  journal = {Computers in Industry}, volume = {162}, pages = {104129}, year = {2024}, doi = {10.1016/j.compind.2024.104129}
}

@article{turner2019decision,
  title   = {Intelligent decision support for maintenance: an overview and future trends},
  author  = {Turner, Chris J. and Emmanouilidis, Christos and Tomiyama, Tetsuo and Tiwari, Ashutosh and Roy, Rajkumar},
  journal = {International Journal of Computer Integrated Manufacturing}, volume = {32}, number = {10}, pages = {936--959}, year = {2019}, doi = {10.1080/0951192X.2019.1667033}
}

@inproceedings{deng2024prescription,
  title     = {From Prediction to Prescription: Large Language Model Agent for Context-Aware Maintenance Decision Support},
  author    = {Deng, Hang and Namoano, Bernadin and Zheng, Bin and Khan, Samir and Erkoyuncu, John Ahmet},
  booktitle = {PHM Society European Conference}, volume = {8}, number = {1}, year = {2024}, doi = {10.36001/phme.2024.v8i1.4114}
}

@inproceedings{vidyaratne2024troubleshooting,
  title     = {Generating Troubleshooting Trees for Industrial Equipment using Large Language Models},
  author    = {Vidyaratne, Lasitha and Lee, Xiao Yee and Kumar, Ahmed and Watanabe, Takanori and Farahat, Ahmed and Gupta, Chetan},
  booktitle = {2024 IEEE International Conference on Prognostics and Health Management (ICPHM)}, pages = {116--125}, year = {2024}, doi = {10.1109/ICPHM61352.2024.10626823}
}

@article{wang2024ontology,
  title   = {Ontology-integrated tuning of large language model for intelligent maintenance},
  author  = {Wang, Peng and Karigiannis, John and Gao, Robert X.},
  journal = {CIRP Annals}, volume = {73}, number = {1}, pages = {361--364}, year = {2024}, doi = {10.1016/j.cirp.2024.04.012}
}

@article{emmanouilidis2019human,
  title   = {Enabling the human in the loop: Linked data and knowledge in industrial cyber-physical systems},
  author  = {Emmanouilidis, Christos and Pistofidis, Petros and Bertoncelj, Luka and Katsouros, Vassilis and Fournaris, Apostolos and Koulamas, Christos and Ruiz-Carcel, Cristobal},
  journal = {Annual Reviews in Control}, volume = {47}, pages = {249--265}, year = {2019}, doi = {10.1016/j.arcontrol.2019.03.004}
}

@inproceedings{panickssery2024self,
  title     = {{LLM} Evaluators Recognize and Favor Their Own Generations},
  author    = {Panickssery, Arjun and Bowman, Samuel R. and Feng, Shi},
  booktitle = {Advances in Neural Information Processing Systems (NeurIPS)}, year = {2024}
}

@misc{laban2025lost,
  title  = {{LLM}s Get Lost in Multi-Turn Conversation},
  author = {Laban, Philippe and Hayashi, Hiroaki and Zhou, Yingbo and Neville, Jennifer},
  year   = {2025}, eprint = {2505.06120}, archivePrefix = {arXiv}, primaryClass = {cs.CL}
}

@misc{openai2024gpt4omini,
  title = {{GPT-4o} mini: Advancing Cost-Efficient Intelligence},
  author = {{OpenAI}},
  year = {2024}, howpublished = {\url{https://openai.com/index/gpt-4o-mini-advancing-cost-efficient-intelligence/}}
}

@misc{anthropic2025sonnet45,
  title  = {{Claude Sonnet 4.5} System Card},
  author = {{Anthropic}},
  year   = {2025},
  howpublished = {\url{https://www.anthropic.com/claude-sonnet-4-5-system-card}}
}

@misc{anthropic2025haiku45,
  title  = {{Claude Haiku 4.5} System Card},
  author = {{Anthropic}},
  year   = {2025},
  howpublished = {\url{https://www.anthropic.com/claude-haiku-4-5-system-card}}
}

@misc{grattafiori2024llama,
      title={The Llama 3 Herd of Models}, 
      author={Aaron Grattafiori and Abhimanyu Dubey and Abhinav Jauhri and Abhinav Pandey and Abhishek Kadian and Ahmad Al-Dahle and Aiesha Letman and Akhil Mathur and Alan Schelten and Alex Vaughan and Amy Yang and Angela Fan and Anirudh Goyal and Anthony Hartshorn and Aobo Yang and Archi Mitra and Archie Sravankumar and Artem Korenev and Arthur Hinsvark and Arun Rao and Aston Zhang and Aurelien Rodriguez and Austen Gregerson and Ava Spataru and Baptiste Roziere and Bethany Biron and Binh Tang and Bobbie Chern and Charlotte Caucheteux and Chaya Nayak and Chloe Bi and Chris Marra and Chris McConnell and Christian Keller and Christophe Touret and Chunyang Wu and Corinne Wong and Cristian Canton Ferrer and Cyrus Nikolaidis and Damien Allonsius and Daniel Song and Danielle Pintz and Danny Livshits and Danny Wyatt and David Esiobu and Dhruv Choudhary and Dhruv Mahajan and Diego Garcia-Olano and Diego Perino and Dieuwke Hupkes and Egor Lakomkin and Ehab AlBadawy and Elina Lobanova and Emily Dinan and Eric Michael Smith and Filip Radenovic and Francisco Guzmán and Frank Zhang and Gabriel Synnaeve and Gabrielle Lee and Georgia Lewis Anderson and Govind Thattai and Graeme Nail and Gregoire Mialon and Guan Pang and Guillem Cucurell and Hailey Nguyen and Hannah Korevaar and Hu Xu and Hugo Touvron and Iliyan Zarov and Imanol Arrieta Ibarra and Isabel Kloumann and Ishan Misra and Ivan Evtimov and Jack Zhang and Jade Copet and Jaewon Lee and Jan Geffert and Jana Vranes and Jason Park and Jay Mahadeokar and Jeet Shah and Jelmer van der Linde and Jennifer Billock and Jenny Hong and Jenya Lee and Jeremy Fu and Jianfeng Chi and Jianyu Huang and Jiawen Liu and Jie Wang and Jiecao Yu and Joanna Bitton and Joe Spisak and Jongsoo Park and Joseph Rocca and Joshua Johnstun and Joshua Saxe and Junteng Jia and Kalyan Vasuden Alwala and Karthik Prasad and Kartikeya Upasani and Kate Plawiak and Ke Li and Kenneth Heafield and Kevin Stone and Khalid El-Arini and Krithika Iyer and Kshitiz Malik and Kuenley Chiu and Kunal Bhalla and Kushal Lakhotia and Lauren Rantala-Yeary and Laurens van der Maaten and Lawrence Chen and Liang Tan and Liz Jenkins and Louis Martin and Lovish Madaan and Lubo Malo and Lukas Blecher and Lukas Landzaat and Luke de Oliveira and Madeline Muzzi and Mahesh Pasupuleti and Mannat Singh and Manohar Paluri and Marcin Kardas and Maria Tsimpoukelli and Mathew Oldham and Mathieu Rita and Maya Pavlova and Melanie Kambadur and Mike Lewis and Min Si and Mitesh Kumar Singh and Mona Hassan and Naman Goyal and Narjes Torabi and Nikolay Bashlykov and Nikolay Bogoychev and Niladri Chatterji and Ning Zhang and Olivier Duchenne and Onur Çelebi and Patrick Alrassy and Pengchuan Zhang and Pengwei Li and Petar Vasic and Peter Weng and Prajjwal Bhargava and Pratik Dubal and Praveen Krishnan and Punit Singh Koura and Puxin Xu and Qing He and Qingxiao Dong and Ragavan Srinivasan and Raj Ganapathy and Ramon Calderer and Ricardo Silveira Cabral and Robert Stojnic and Roberta Raileanu and Rohan Maheswari and Rohit Girdhar and Rohit Patel and Romain Sauvestre and Ronnie Polidoro and Roshan Sumbaly and Ross Taylor and Ruan Silva and Rui Hou and Rui Wang and Saghar Hosseini and Sahana Chennabasappa and Sanjay Singh and Sean Bell and Seohyun Sonia Kim and Sergey Edunov and Shaoliang Nie and Sharan Narang and Sharath Raparthy and Sheng Shen and Shengye Wan and Shruti Bhosale and Shun Zhang and Simon Vandenhende and Soumya Batra and Spencer Whitman and Sten Sootla and Stephane Collot and Suchin Gururangan and Sydney Borodinsky and Tamar Herman and Tara Fowler and Tarek Sheasha and Thomas Georgiou and Thomas Scialom and Tobias Speckbacher and Todor Mihaylov and Tong Xiao and Ujjwal Karn and Vedanuj Goswami and Vibhor Gupta and Vignesh Ramanathan and Viktor Kerkez and Vincent Gonguet and Virginie Do and Vish Vogeti and Vítor Albiero and Vladan Petrovic and Weiwei Chu and Wenhan Xiong and Wenyin Fu and Whitney Meers and Xavier Martinet and Xiaodong Wang and Xiaofang Wang and Xiaoqing Ellen Tan and Xide Xia and Xinfeng Xie and Xuchao Jia and Xuewei Wang and Yaelle Goldschlag and Yashesh Gaur and Yasmine Babaei and Yi Wen and Yiwen Song and Yuchen Zhang and Yue Li and Yuning Mao and Zacharie Delpierre Coudert and Zheng Yan and Zhengxing Chen and Zoe Papakipos and Aaditya Singh and Aayushi Srivastava and Abha Jain and Adam Kelsey and Adam Shajnfeld and Adithya Gangidi and Adolfo Victoria and Ahuva Goldstand and Ajay Menon and Ajay Sharma and Alex Boesenberg and Alexei Baevski and Allie Feinstein and Amanda Kallet and Amit Sangani and Amos Teo and Anam Yunus and Andrei Lupu and Andres Alvarado and Andrew Caples and Andrew Gu and Andrew Ho and Andrew Poulton and Andrew Ryan and Ankit Ramchandani and Annie Dong and Annie Franco and Anuj Goyal and Aparajita Saraf and Arkabandhu Chowdhury and Ashley Gabriel and Ashwin Bharambe and Assaf Eisenman and Azadeh Yazdan and Beau James and Ben Maurer and Benjamin Leonhardi and Bernie Huang and Beth Loyd and Beto De Paola and Bhargavi Paranjape and Bing Liu and Bo Wu and Boyu Ni and Braden Hancock and Bram Wasti and Brandon Spence and Brani Stojkovic and Brian Gamido and Britt Montalvo and Carl Parker and Carly Burton and Catalina Mejia and Ce Liu and Changhan Wang and Changkyu Kim and Chao Zhou and Chester Hu and Ching-Hsiang Chu and Chris Cai and Chris Tindal and Christoph Feichtenhofer and Cynthia Gao and Damon Civin and Dana Beaty and Daniel Kreymer and Daniel Li and David Adkins and David Xu and Davide Testuggine and Delia David and Devi Parikh and Diana Liskovich and Didem Foss and Dingkang Wang and Duc Le and Dustin Holland and Edward Dowling and Eissa Jamil and Elaine Montgomery and Eleonora Presani and Emily Hahn and Emily Wood and Eric-Tuan Le and Erik Brinkman and Esteban Arcaute and Evan Dunbar and Evan Smothers and Fei Sun and Felix Kreuk and Feng Tian and Filippos Kokkinos and Firat Ozgenel and Francesco Caggioni and Frank Kanayet and Frank Seide and Gabriela Medina Florez and Gabriella Schwarz and Gada Badeer and Georgia Swee and Gil Halpern and Grant Herman and Grigory Sizov and Guangyi and Zhang and Guna Lakshminarayanan and Hakan Inan and Hamid Shojanazeri and Han Zou and Hannah Wang and Hanwen Zha and Haroun Habeeb and Harrison Rudolph and Helen Suk and Henry Aspegren and Hunter Goldman and Hongyuan Zhan and Ibrahim Damlaj and Igor Molybog and Igor Tufanov and Ilias Leontiadis and Irina-Elena Veliche and Itai Gat and Jake Weissman and James Geboski and James Kohli and Janice Lam and Japhet Asher and Jean-Baptiste Gaya and Jeff Marcus and Jeff Tang and Jennifer Chan and Jenny Zhen and Jeremy Reizenstein and Jeremy Teboul and Jessica Zhong and Jian Jin and Jingyi Yang and Joe Cummings and Jon Carvill and Jon Shepard and Jonathan McPhie and Jonathan Torres and Josh Ginsburg and Junjie Wang and Kai Wu and Kam Hou U and Karan Saxena and Kartikay Khandelwal and Katayoun Zand and Kathy Matosich and Kaushik Veeraraghavan and Kelly Michelena and Keqian Li and Kiran Jagadeesh and Kun Huang and Kunal Chawla and Kyle Huang and Lailin Chen and Lakshya Garg and Lavender A and Leandro Silva and Lee Bell and Lei Zhang and Liangpeng Guo and Licheng Yu and Liron Moshkovich and Luca Wehrstedt and Madian Khabsa and Manav Avalani and Manish Bhatt and Martynas Mankus and Matan Hasson and Matthew Lennie and Matthias Reso and Maxim Groshev and Maxim Naumov and Maya Lathi and Meghan Keneally and Miao Liu and Michael L. Seltzer and Michal Valko and Michelle Restrepo and Mihir Patel and Mik Vyatskov and Mikayel Samvelyan and Mike Clark and Mike Macey and Mike Wang and Miquel Jubert Hermoso and Mo Metanat and Mohammad Rastegari and Munish Bansal and Nandhini Santhanam and Natascha Parks and Natasha White and Navyata Bawa and Nayan Singhal and Nick Egebo and Nicolas Usunier and Nikhil Mehta and Nikolay Pavlovich Laptev and Ning Dong and Norman Cheng and Oleg Chernoguz and Olivia Hart and Omkar Salpekar and Ozlem Kalinli and Parkin Kent and Parth Parekh and Paul Saab and Pavan Balaji and Pedro Rittner and Philip Bontrager and Pierre Roux and Piotr Dollar and Polina Zvyagina and Prashant Ratanchandani and Pritish Yuvraj and Qian Liang and Rachad Alao and Rachel Rodriguez and Rafi Ayub and Raghotham Murthy and Raghu Nayani and Rahul Mitra and Rangaprabhu Parthasarathy and Raymond Li and Rebekkah Hogan and Robin Battey and Rocky Wang and Russ Howes and Ruty Rinott and Sachin Mehta and Sachin Siby and Sai Jayesh Bondu and Samyak Datta and Sara Chugh and Sara Hunt and Sargun Dhillon and Sasha Sidorov and Satadru Pan and Saurabh Mahajan and Saurabh Verma and Seiji Yamamoto and Sharadh Ramaswamy and Shaun Lindsay and Shaun Lindsay and Sheng Feng and Shenghao Lin and Shengxin Cindy Zha and Shishir Patil and Shiva Shankar and Shuqiang Zhang and Shuqiang Zhang and Sinong Wang and Sneha Agarwal and Soji Sajuyigbe and Soumith Chintala and Stephanie Max and Stephen Chen and Steve Kehoe and Steve Satterfield and Sudarshan Govindaprasad and Sumit Gupta and Summer Deng and Sungmin Cho and Sunny Virk and Suraj Subramanian and Sy Choudhury and Sydney Goldman and Tal Remez and Tamar Glaser and Tamara Best and Thilo Koehler and Thomas Robinson and Tianhe Li and Tianjun Zhang and Tim Matthews and Timothy Chou and Tzook Shaked and Varun Vontimitta and Victoria Ajayi and Victoria Montanez and Vijai Mohan and Vinay Satish Kumar and Vishal Mangla and Vlad Ionescu and Vlad Poenaru and Vlad Tiberiu Mihailescu and Vladimir Ivanov and Wei Li and Wenchen Wang and Wenwen Jiang and Wes Bouaziz and Will Constable and Xiaocheng Tang and Xiaojian Wu and Xiaolan Wang and Xilun Wu and Xinbo Gao and Yaniv Kleinman and Yanjun Chen and Ye Hu and Ye Jia and Ye Qi and Yenda Li and Yilin Zhang and Ying Zhang and Yossi Adi and Youngjin Nam and Yu and Wang and Yu Zhao and Yuchen Hao and Yundi Qian and Yunlu Li and Yuzi He and Zach Rait and Zachary DeVito and Zef Rosnbrick and Zhaoduo Wen and Zhenyu Yang and Zhiwei Zhao and Zhiyu Ma},
      year={2024},
      eprint={2407.21783},
      archivePrefix={arXiv},
      primaryClass={cs.AI},
      url={https://arxiv.org/abs/2407.21783}, 
}

@misc{yang2025qwen3,
      title={Qwen3 Technical Report}, 
      author={An Yang and Anfeng Li and Baosong Yang and Beichen Zhang and Binyuan Hui and Bo Zheng and Bowen Yu and Chang Gao and Chengen Huang and Chenxu Lv and Chujie Zheng and Dayiheng Liu and Fan Zhou and Fei Huang and Feng Hu and Hao Ge and Haoran Wei and Huan Lin and Jialong Tang and Jian Yang and Jianhong Tu and Jianwei Zhang and Jianxin Yang and Jiaxi Yang and Jing Zhou and Jingren Zhou and Junyang Lin and Kai Dang and Keqin Bao and Kexin Yang and Le Yu and Lianghao Deng and Mei Li and Mingfeng Xue and Mingze Li and Pei Zhang and Peng Wang and Qin Zhu and Rui Men and Ruize Gao and Shixuan Liu and Shuang Luo and Tianhao Li and Tianyi Tang and Wenbiao Yin and Xingzhang Ren and Xinyu Wang and Xinyu Zhang and Xuancheng Ren and Yang Fan and Yang Su and Yichang Zhang and Yinger Zhang and Yu Wan and Yuqiong Liu and Zekun Wang and Zeyu Cui and Zhenru Zhang and Zhipeng Zhou and Zihan Qiu},
      year={2025},
      eprint={2505.09388},
      archivePrefix={arXiv},
      primaryClass={cs.CL},
      url={https://arxiv.org/abs/2505.09388}, 
}

@misc{mistral2025small3,
  title = {Mistral Small 3},
  author = {{Mistral AI Team}},
  year = {2025}, howpublished = {\url{https://mistral.ai/news/mistral-small-3/}}
}

@misc{openai2025gptoss,
      title={gpt-oss-120b \& gpt-oss-20b Model Card}, 
      author={Sandhini  Agarwal and Lama Ahmad and Jason Ai and Sam Altman and Andy Applebaum and Edwin Arbus and Rahul K. Arora and Yu Bai and Bowen Baker and Haiming Bao and Boaz Barak and Ally Bennett and Tyler Bertao and Nivedita Brett and Eugene Brevdo and Greg Brockman and Sebastien Bubeck and Che Chang and Kai Chen and Mark Chen and Enoch Cheung and Aidan Clark and Dan Cook and Marat Dukhan and Casey Dvorak and Kevin Fives and Vlad Fomenko and Timur Garipov and Kristian Georgiev and Mia Glaese and Tarun Gogineni and Adam Goucher and Lukas Gross and Katia Gil Guzman and John Hallman and Jackie Hehir and Johannes Heidecke and Alec Helyar and Haitang Hu and Romain Huet and Jacob Huh and Saachi Jain and Zach Johnson and Chris Koch and Irina Kofman and Dominik Kundel and Jason Kwon and Volodymyr Kyrylov and Elaine Ya Le and Guillaume Leclerc and James Park Lennon and Scott Lessans and Mario Lezcano-Casado and Yuanzhi Li and Zhuohan Li and Ji Lin and Jordan Liss and Lily and Liu and Jiancheng Liu and Kevin Lu and Chris Lu and Zoran Martinovic and Lindsay McCallum and Josh McGrath and Scott McKinney and Aidan McLaughlin and Song Mei and Steve Mostovoy and Tong Mu and Gideon Myles and Alexander Neitz and Alex Nichol and Jakub Pachocki and Alex Paino and Dana Palmie and Ashley Pantuliano and Giambattista Parascandolo and Jongsoo Park and Leher Pathak and Carolina Paz and Ludovic Peran and Dmitry Pimenov and Michelle Pokrass and Elizabeth Proehl and Huida Qiu and Gaby Raila and Filippo Raso and Hongyu Ren and Kimmy Richardson and David Robinson and Bob Rotsted and Hadi Salman and Suvansh Sanjeev and Max Schwarzer and D. Sculley and Harshit Sikchi and Kendal Simon and Karan Singhal and Yang Song and Dane Stuckey and Zhiqing Sun and Philippe Tillet and Sam Toizer and Foivos Tsimpourlas and Nikhil Vyas and Eric Wallace and Xin Wang and Miles Wang and Olivia Watkins and Kevin Weil and Amy Wendling and Kevin Whinnery and Cedric Whitney and Hannah Wong and Lin Yang and Yu Yang and Michihiro Yasunaga and Kristen Ying and Wojciech Zaremba and Wenting Zhan and Cyril Zhang and Brian Zhang and Eddie Zhang and Shengjia Zhao},
      year={2025},
      eprint={2508.10925},
      archivePrefix={arXiv},
      primaryClass={cs.CL},
      url={https://arxiv.org/abs/2508.10925}, 
}

@misc{nvidia2026nemotron,
  title = {Nemotron 3 Super: An Open Hybrid {Mamba}-Transformer {MoE} for Agentic Reasoning},
  author = {{NVIDIA}},
  year = {2026}, howpublished = {\url{https://research.nvidia.com/labs/nemotron/Nemotron-3-Super}}
}

@misc{meta2025llama4,
  title = {The {Llama} 4 herd: The beginning of a new era of natively multimodal {AI} innovation},
  author = {{Llama Team}},
  year = {2025},
  howpublished = {\url{https://ai.meta.com/blog/llama-4-multimodal-intelligence/}}
}

@misc{openrouter2024,
  title = {{OpenRouter}: A Unified Interface for Large Language Models},
  author = {{OpenRouter}},
  year = {2024},
  howpublished = {\url{https://openrouter.ai}}
}

@misc{streamlit,
  title = {{Streamlit}: A Faster Way to Build and Share Data Apps},
  author = {{Streamlit Inc.}},
  year = {2019},
  howpublished = {\url{https://streamlit.io}}
}

@inproceedings{poesia2022synchromesh,
  title     = {Synchromesh: Reliable code generation from pre-trained language models},
  author    = {Poesia, Gabriel and Polozov, Oleksandr and Le, Vu and Tiwari, Ashish and Soares, Gustavo and Meek, Christopher and Gulwani, Sumit},
  booktitle = {International Conference on Learning Representations (ICLR)},
  year      = {2022}
}

@inproceedings{ravichander2019privacyqa,
  title = {Question Answering for Privacy Policies: Combining Computational and Legal Perspectives},
  author = {Ravichander, Abhilasha and Black, Alan W. and Wilson, Shomir and Norton, Thomas and Sadeh, Norman},
  booktitle = {Proceedings of the 2019 Conference on Empirical Methods in Natural Language Processing and the 9th International Joint Conference on Natural Language Processing (EMNLP-IJCNLP)},
  pages = {4947--4958},
  year = {2019}
}

@article{brundage2021tlp,
  title   = {Technical language processing: Unlocking maintenance knowledge},
  author  = {Brundage, Michael P. and Sexton, Thurston and Hodkiewicz, Melinda and Dima, Alden and Lukens, Sarah},
  journal = {Manufacturing Letters}, volume = {27}, pages = {42--46}, year = {2021},
  doi = {10.1016/j.mfglet.2020.11.001}
}

@article{dima2021adapting,
  title   = {Adapting natural language processing for technical text},
  author  = {Dima, Alden and Lukens, Sarah and Hodkiewicz, Melinda and Sexton, Thurston and Brundage, Michael P.},
  journal = {Applied AI Letters}, volume = {2}, number = {3}, pages = {e33}, year = {2021},
  doi = {10.1002/ail2.33}
}

@techreport{iso5807,
  author      = {{International Organization for Standardization}},
  title       = {Information processing -- Documentation symbols and conventions for data, program and system flowcharts, program network charts and system resources charts},
  number      = {ISO 5807:1985},
  institution = {International Organization for Standardization},
  address     = {Geneva, Switzerland},
  year        = {1985}
}

\clearpage
\appendix

\section{Notation}
\label{app:notation}

Table~\ref{tab:notation} collects the graph and scoring symbols used throughout the paper, in the order they are introduced in \S\ref{sec:task}.

\begin{table}[!ht]
\centering
\small
\setlength{\tabcolsep}{6pt}
\renewcommand{\arraystretch}{1.15}
\begin{tabular}{ll}
\toprule
\textbf{Symbol} & \textbf{Definition} \\
\midrule
$G=(V,E)$             & Directed procedure graph \\
$v$                   & Node (diagnostic step) \\
$e=(v\rightarrow v')$ & Directed edge (admissible transition) \\
$\ell(e)$             & Observation labelling edge $e$ \\
$L(v)$                & Admissible observation labels at $v$ \\
$v_t$                 & Current node at turn $t$ \\
$v^{*}_t$     & Successor of $v_t$ entailed by $o_t$ \\
$o_t$                 & Operator utterance at turn $t$ \\
$h_t$                 & Conversation history up to turn $t$ \\
$\hat{a}_t$           & Predicted next action at turn $t$ \\
$\rho(\cdot)$ & Node a response resolves to, or $\varnothing$ \\
$\tau$                & Jaccard threshold for node matching \\
$\tau^{*}$    & Calibrated value of $\tau$ \\
$\kappa$              & Cohen's $\kappa$ (agreement) \\
\bottomrule
\end{tabular}
\caption{Graph and scoring symbols.}
\label{tab:notation}
\end{table}

\section{Dataset Statistics and Path Enumeration}
\label{app:dataset}

Table~\ref{tab:dataset_stats} gives aggregate node-type statistics, with graphs grouped into descriptive size bands by node count and nodes grouped by type following \citet{iso5807}. Table~\ref{tab:per_graph_stats} lists per-graph topology for all fifty flowcharts, enabling verification that a reproduced dataset matches ours. 

Anonymisation proceeds in three steps. (1) A model first extracts proprietary identifiers. (2) A second model rewrites each procedure into an abstract generic industrial domain. (3) Subject-matter experts at the partner organisation carefully review and confirm that temporal dependencies, nodes, branching and escalation logic survived the anonymisation while maintaining topology.

\begin{table}[htbp]
\centering
\small
\begin{tabular}{lrlr}
\toprule
\textbf{Topology size} & \textbf{Count} & \textbf{Node type} & \textbf{Count} \\
\midrule
Small ($\le 14$)  & 6  & Decision   & 421 \\
Medium (15--30)   & 21 & Process    & 709 \\
Large ($> 30$)    & 23 & Document   & 258 \\
                  &    & Terminator & 179 \\
\midrule
\textbf{Total}    & \textbf{50} & \textbf{Total nodes} & \textbf{1{,}567} \\
\bottomrule
\end{tabular}
\caption{Aggregate statistics across the fifty flowcharts.}
\label{tab:dataset_stats}
\end{table}

\begin{figure*}[t]
 \centering
 \includegraphics[width=\linewidth]{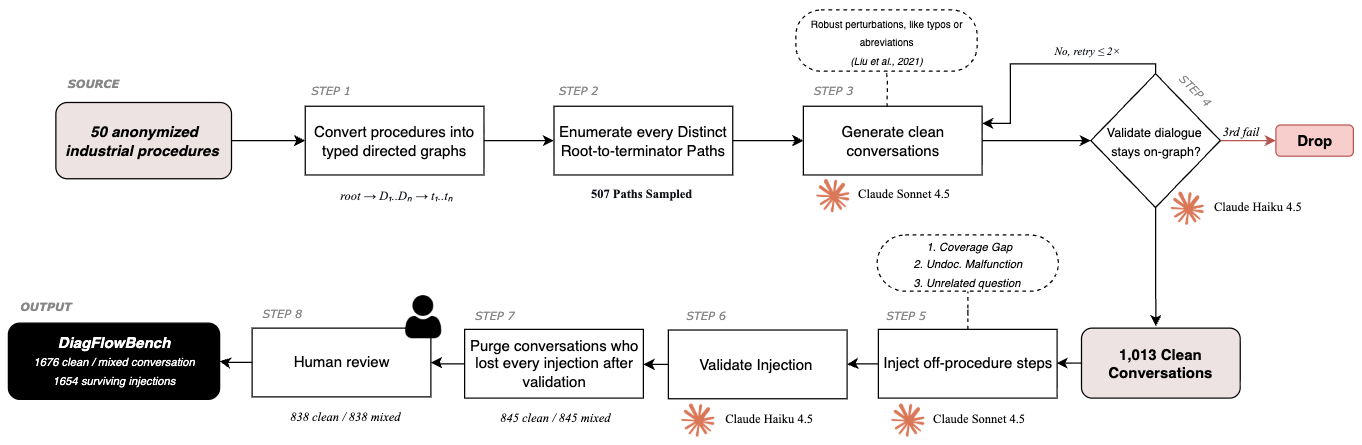}
 \caption{Generation pipeline. Source procedures are anonymised under expert validation, root-to-terminal paths enumerated under greedy set cover, per-path conversations generated and noise-perturbed into 838 cooperative conversations, then 1,654 verified off-procedure injections inserted to form the 838 mixed conversations.}
 \label{fig:generation_pipeline}
\end{figure*}

\begin{table*}[tp]
\centering
\small
\begin{tabular}{lrrrrr lrrrrr}
\toprule
\textbf{ID} & $|V|$ & $|E|$ & \textbf{Dec} & \textbf{Paths} & \textbf{Conv} &
\textbf{ID} & $|V|$ & $|E|$ & \textbf{Dec} & \textbf{Paths} & \textbf{Conv} \\
\midrule
G01 & 14 & 13 & 4 & 4 & 8 & G26 & 23 & 28 & 7 & 9 & 18 \\
G02 &  7 &  6 & 0 & 1 & 2 & G27 & 23 & 28 & 7 & 9 & 18 \\
G03 & 13 & 12 & 4 & 5 & 10 & G28 & 24 & 29 & 7 & 9 & 18 \\
G04 &  9 &  8 & 2 & 3 & 6 & G29 & 26 & 32 & 8 & 10 & 20 \\
G05 & 11 & 10 & 3 & 4 & 8 & G30 & 23 & 28 & 7 & 9 & 18 \\
G06 & 12 & 11 & 3 & 4 & 8 & G31 & 23 & 27 & 6 & 8 & 16 \\
G07 & 19 & 18 & 6 & 7 & 14 & G32 & 23 & 29 & 6 & 8 & 16 \\
G08 & 37 & 36 & 9 & 10 & 20 & G33 & 25 & 29 & 7 & 9 & 18 \\
G09 & 39 & 38 & 10 & 12 & 24 & G34 & 24 & 28 & 6 & 8 & 16 \\
G10 & 56 & 55 & 12 & 14 & 28 & G35 & 28 & 33 & 9 & 11 & 22 \\
G11 & 60 & 59 & 21 & 22 & 44 & G36 & 34 & 40 & 10 & 12 & 24 \\
G12 & 42 & 41 & 12 & 13 & 26 & G37 & 37 & 43 & 10 & 12 & 24 \\
G13 & 21 & 20 & 6 & 7 & 14 & G38 & 39 & 46 & 11 & 13 & 26 \\
G14 & 53 & 52 & 10 & 12 & 24 & G39 & 35 & 41 & 10 & 12 & 24 \\
G15 & 38 & 37 & 9 & 11 & 22 & G40 & 36 & 43 & 10 & 12 & 24 \\
G16 & 41 & 40 & 9 & 11 & 22 & G41 & 50 & 58 & 14 & 16 & 32 \\
G17 & 43 & 43 & 9 & 11 & 22 & G42 & 54 & 64 & 15 & 17 & 34 \\
G18 & 38 & 37 & 9 & 11 & 22 & G43 & 55 & 64 & 16 & 18 & 36 \\
G19 & 35 & 34 & 9 & 9 & 18 & G44 & 49 & 58 & 13 & 15 & 30 \\
G20 & 24 & 27 & 6 & 8 & 16 & G45 & 24 & 29 & 7 & 9 & 18 \\
G21 & 24 & 28 & 6 & 8 & 16 & G46 & 28 & 31 & 7 & 9 & 18 \\
G22 & 23 & 28 & 7 & 9 & 18 & G47 & 26 & 31 & 7 & 9 & 18 \\
G23 & 23 & 27 & 7 & 9 & 18 & G48 & 35 & 41 & 10 & 12 & 24 \\
G24 & 23 & 27 & 6 & 8 & 16 & G49 & 39 & 44 & 10 & 12 & 24 \\
G25 & 23 & 28 & 7 & 9 & 18 & G50 & 56 & 64 & 15 & 17 & 34 \\
\bottomrule
\end{tabular}
\caption{Per-graph topology statistics. $|V|$: nodes, $|E|$: edges, Dec: decision nodes, Paths: root-to-terminal paths selected by the greedy set-cover, Conv: cooperative conversations after $\times$2 noise-perturbation expansion.}
\label{tab:per_graph_stats}
\end{table*}

\subsection{Path Enumeration}
\label{app:paths}

Conversations are seeded from paths through each graph, ensuring full coverage of branches and terminal outcomes without over-sampling large graphs. We frame this as a set-cover problem over decision edges and terminal types, greedily adding root-to-terminal paths by marginal coverage. To prevent large graphs from dominating, the number of paths per graph is capped at $\max(2,\, N_{\text{dec}} + 2)$, where $N_{\text{dec}}$ is the number of decision nodes in the graph. Each selected path yields one cooperative conversation plus one noise-perturbed variant (the robustness pass of \S\ref{sec:construction}), so a graph contributing $p$ paths contributes $2p$ cooperative conversations, and an equal number of mixed conversations after injection. This procedure yields 507 enumerated paths and 1{,}013 generated conversation pairs (accounting for one failed generation). Off-procedure injections are inserted at non-terminal positions in the mixed conversations. The injection count follows a rate sampled uniformly from [0.05, 0.30] of available positions (averaging two, ranging from one to seven), with categories assigned cyclically. Verification and two pair-level filters then reduce the initial 1{,}013 pairs to 838. First, the zero-temperature verifier discards injections entailing an outgoing edge, reducing 2{,}110 planned injections to 1{,}673 and dropping 168 conversations lacking any valid injection (leaving 845 pairs). Second, two PhD annotators supervised by a senior author reviewed the remainder for naturalness. They specifically targeted and removed 7 conversations containing 19 unnatural, self-signalling utterances that felt uncharacteristic of human technicians, yielding the final 838 pairs.

\section{Generation and Verification}
\label{app:gen_prompts}

Figure~\ref{fig:generation_pipeline} illustrates the end-to-end pipeline. All three stages use \texttt{claude-sonnet-4-5} \citep{anthropic2025sonnet45}, selected for its reliability with long structured prompts and instruction-following. Because the generator output is constrained by the verification gate, construction does not heavily depend on this specific choice of model.   

On the first pass, the generator receives the graph and a single root-to-terminal path, producing one operator utterance per node along that path. This pass runs at temperature 0.7, which is typical for open-ended generation.

\begin{promptbox}[Pass 1: Generator System Prompt]
You are generating scripted operator turns for a diagnostic troubleshooting benchmark. You are given a directed graph representing a diagnostic procedure and a specific root-to-terminal path through it. 
For each node on the path, generate an OPERATOR utterance, what a factory operator would say to report the observation or action described by that node. It must be consistent with the edge label that led to this node, sound like natural speech from a maintenance technician, not mention or narrate downstream steps, not refer to node IDs or graph metadata, and be 1 to 3 sentences. 

For \textbf{PROCESS} nodes the operator describes performing the action or its result. For 
\textbf{DECISION} nodes the operator reports the observation matching the outgoing edge label. For 
\textbf{DOCUMENT} nodes the operator references consulting the document.  
For \textbf{TERMINATOR} nodes the operator acknowledges the endpoint. 
Return a JSON array, one object per turn.
\end{promptbox}

The verifier audits every generated script against the graph and the path that produced it. To maximise determinism, the verification process operates at temperature zero. Rather than discarding invalid scripts outright, the verifier returns itemised corrections, allowing the generation prompts to be amended iteratively. 

\begin{promptboxnb}[Pass 2: Verifier System Prompt (zero temperature)]
You are a quality reviewer for a diagnostic troubleshooting benchmark. You are given a directed graph, a path through it, and a generated conversation. Verify each operator utterance and flag:  \textbf{EDGE INCONSISTENCY} (does not match the edge label for this branch), \textbf{STEP LEAKAGE} (implies knowledge of steps not yet reached), \textbf{UNNATURALNESS} (robotic or uses graph terminology), and \textbf{INFORMATION LEAK} (reveals graph structure, node IDs, or metadata). For each flagged utterance provide a corrected version. Return JSON with the verified status, the issue list, and the corrected turns.
\end{promptboxnb}

Each verified script is then duplicated and rewritten to carry the surface characteristics of maintenance work-order text. This pass alters wording alone, leaving the path through the graph and the observation reported at each turn
unchanged.

\begin{promptbox}[Noise-Perturbation Robustness Pass]
Rewrite operator utterances to mimic how a technician types quickly on a tablet or radio interface. 

Apply a mix of the following, not all at once, but enough to feel clearly informal:  \textbf{Lowercase} or no opening capitalisation,  \textbf{Missing punctuation}, 2--3 realistic typos (transposed or dropped letters),  \textbf{Telegraphic phrasing} (`insertion depth ok' instead of `the insertion depth is within specification'),  \textbf{Abbreviated units and terms} (`temp', `spec', `engmt force', `dim deviation'),  \textbf{Filler words} (`yeah', `ok so', `looks like'),  \textbf{Occasional run-on sentences} without commas. 
The utterance must remain clearly understandable and semantically identical to the original.
\end{promptbox}

\subsection{Off-procedure injection prompts}

The off-procedure injection utterances are generated and verified with a generator and verifier pair mirroring the one used earlier. Both use the same instruction-following \texttt{claude-sonnet-4-5} model \citep{anthropic2025sonnet45}.

\begin{promptbox}[Injection Generator System Prompt]
You are generating a single off-procedure operator utterance for a diagnostic troubleshooting benchmark. 
The operator is currently at a specific node in a diagnostic procedure. You must generate one operator turn that reports something the procedure does NOT handle.

You will be given: current\_node\_text (what the procedure is currently asking or instructing), outgoing\_edge\_labels (the only observations that are valid at this node), and failure\_category (what kind of off-procedure input to generate).
Failure categories: \textbf{coverage\_gap}, a symptom or observation the procedure does not address; the operator notices something real about the equipment but the algorithm has no branch for it. 

\textbf{undocumented\_malfunction}, a specific fault condition not listed as any branch condition at this node; the operator names a specific part or reading the procedure does not cover.

\textbf{unrelated\_question}, a question entirely unrelated to the current diagnostic step (e.g.\ about a different system, safety protocol, scheduling, or documentation).

The generated utterance MUST sound like natural speech from a maintenance technician (1--2 sentences), NOT entail or imply any of the outgoing\_edge\_labels listed, NOT refer to node IDs or graph metadata, and be plausible given the industrial maintenance context. Return JSON: \{``utterance'': ``\ldots''\}
\end{promptbox}

\begin{promptbox}[Injection Verifier System Prompt (zero temperature)]
You are verifying that an operator utterance does NOT entail any outgoing edge of the current node in a diagnostic procedure.
You will be given: current\_node\_text (what the procedure is asking at this node), outgoing\_edge\_labels (the only valid observations at this node), and utterance (the turn to verify).
An utterance ``entails'' an edge label if a reasonable person reading it would conclude that the operator is reporting that specific observation or confirming that specific condition. Paraphrases, synonyms, and clear implications count. Return JSON: \{``valid'': true/false, ``reason'': ``\ldots''\}
\end{promptbox}

\section{Evaluation}
\label{app:eval_prompts}

\begin{figure*}[t]
 \centering
 \includegraphics[width=\linewidth]{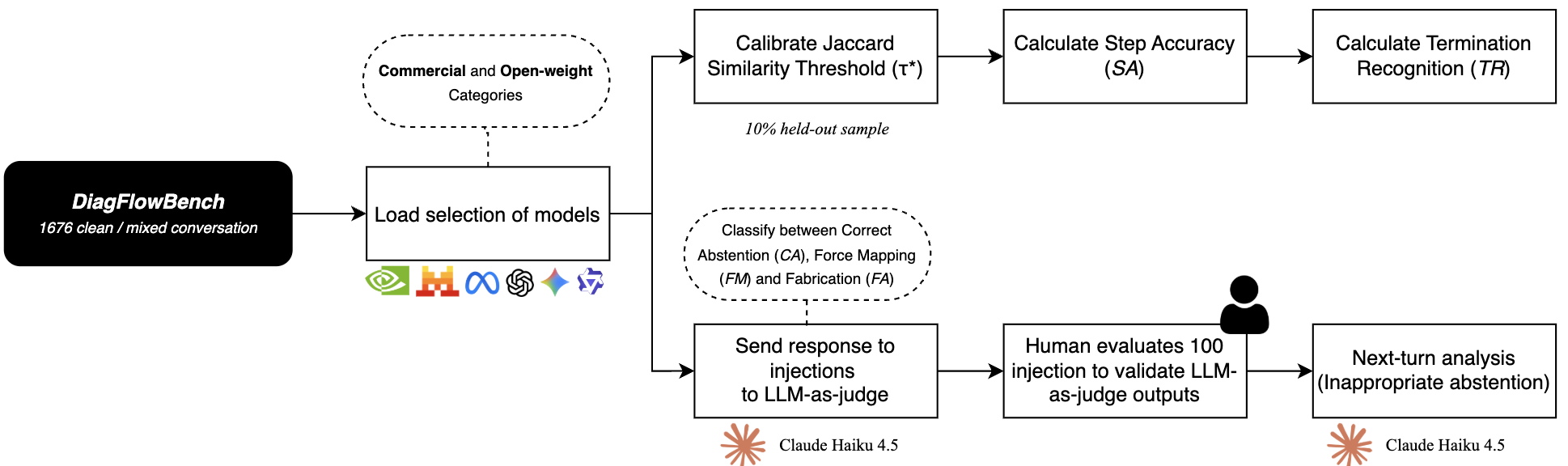}
 \caption{Evaluation pipeline. Cooperative conversations are scored by threshold-calibrated Jaccard matching for step accuracy and termination recognition. Injected turns in mixed conversations are classified by the judge into FA, FM, or CA, with a second pass measuring inappropriate abstention on the following on-procedure turn.}
 \label{fig:evaluation_pipeline}
\end{figure*}

Figure~\ref{fig:evaluation_pipeline} shows how a model is scored, and Appendix~\ref{app:contamination} covers the second pass on the turn that follows an injection. The evaluation prompt is reproduced below.

\begin{promptbox}[Evaluation Prompt]
You are a diagnostic troubleshooting assistant for industrial maintenance. An operator is reporting observations from inspecting faulty equipment. Suggest the NEXT diagnostic action the operator should take, based on the conversation history and the procedure graph below. PROCEDURE GRAPH (JSON): \{graph\_json\} 
Rules: Suggest exactly ONE next diagnostic action per turn. Your suggestion MUST correspond to a node in the procedure graph above. Follow the edges of the graph based on the operator's reported observations. If the operator's observation does not correspond to any node in the procedure graph, do not suggest a procedure step. If the procedure has reached a terminator node, say so clearly and do not suggest further actions. Keep your response to 1 to 3 sentences.
\end{promptbox}

\subsection{Choice-framed abstention ablation}
\label{app:counterfactual}

In deployed advisory systems, the instruction to abstain typically competes against other task objectives. The baseline evaluation prompt (\S\ref{sec:prompting}) preserves this tension. However, to isolate the intrinsic rate of FM, we introduce an ablation that removes the obligation to select a procedure step. We re-evaluated every off-procedure turn under the choice-framed prompt below, restricting the instruction set entirely to the abstention decision. By requiring the model to verify coverage before naming an action, abstention detection becomes the primary objective. The FM surviving this framing therefore represents a rigorous lower bound rather than a prompt artefact. We measure this residual on Llama~3.3~70B, Nemotron~3~Super, and Gemini~2.5~Flash (the three models with the highest CA rates in Table~\ref{tab:main_results}) to establish the bound where the propensity to abstain is already strongest.

\begin{promptbox}[Choice-framed Prompt]
You are a diagnostic troubleshooting assistant for industrial maintenance. An operator is reporting observations from inspecting faulty equipment. Your task is to suggest the NEXT diagnostic action the operator should take, based on the conversation history and the procedure graph below, or to ABSTAIN if there is none.

PROCEDURE GRAPH (JSON): \{graph\_json\}

Rules:\\
- \textbf{SUGGEST A PROCEDURE STEP}: If the operator's observation corresponds to a documented step in the procedure graph, name exactly one next diagnostic action that appears as a node in the graph above. Follow the graph edges based on what the operator has reported. Your suggestion must be a real node from the graph.\\
- \textbf{ABSTAIN}: If the operator's observation does not correspond to any node in the procedure graph (for instance, it describes a symptom, condition, or question that the documented procedure does not address) clearly say so in plain language. Briefly explain that the observation falls outside the procedure and do not suggest a procedure node.\\
- If the procedure has reached a terminator node, say so clearly and do not suggest further actions.\\
- Keep your response concise (1-3 sentences).
\end{promptbox}

\begin{table}[htbp]
\centering
\small
\setlength{\tabcolsep}{4pt}
\begin{tabular}{lrrrrr}
\toprule
\textbf{Model} & \textbf{FM} & \textbf{FM$_{c}$} & \textbf{$\Delta$FM} & \textbf{CA} & \textbf{CA$_{c}$} \\
\midrule
Llama 3.3 70B    & 15.7 &  9.0 & $-6.7$  & 81.3 & 89.1 \\
Nemotron 3 Super & 22.1 & 11.6 & $-10.5$ & 73.9 & 87.3 \\
Gemini 2.5 Flash & 25.6 & 11.4 & $-14.2$ & 72.0 & 87.0 \\
\bottomrule
\end{tabular}
\caption{Off-procedure outcomes under the original evaluation prompt and under the choice-framed prompt (subscript $c$), all values in \%.}
\label{tab:counterfactual}
\end{table}

Table~\ref{tab:counterfactual} reports the comparison. FM falls by 6.7 to 14.2 points and CA rises to between 87.0 and 89.1\%, indicating that a substantial portion of the baseline error stems from instruction competition. However, a residual FM rate between 9.0 and 11.6\% persists despite this explicit framing. This residual remains present even in Llama~3.3~70B (the model with the highest baseline CA), confirming an intrinsic bias toward task resolution rather than a pure prompt artefact. Because the choice-framed ablation isolates the abstention decision in a controlled setting rather than a viable deployment prompt, this observed reduction establishes the practical limit of what prompt engineering alone can mitigate.

\subsection{Models and Hyperparameters}
\label{app:models}

\begin{table*}[tp]
\centering
\small
\begin{tabular}{lllll}
\toprule
\textbf{Model} & \textbf{Provider} & \textbf{Params} & \textbf{Architecture} & \textbf{Category} \\
\midrule
Gemini 2.5 Flash      & Google     & undisclosed  & instruct, proprietary & Commercial \\
GPT-4o Mini           & OpenAI     & undisclosed  & instruct, proprietary & Commercial \\
Llama 4 Scout         & Meta       & 109B total, 17B active & MoE instruct & Open (Llama family) \\
Llama 4 Maverick      & Meta       & 400B total, 17B active & MoE instruct & Open (Llama family) \\
Llama 3.3 70B         & Meta       & 70B          & dense instruct        & Open (Llama family) \\
Qwen3 235B Thinking   & Alibaba    & 235B (A22B)  & MoE, reasoning        & Open \\
Qwen3 30B Thinking    & Alibaba    & 30B (A3B)    & MoE, reasoning        & Open \\
Mistral Small 24B     & Mistral AI & 24B          & dense instruct        & Open \\
GPT-OSS 120B          & OpenAI     & 120B (A5.1B) & MoE                   & Open \\
Nemotron 3 Super 120B & NVIDIA     & 120B (A12B)  & MoE                   & Open \\
\bottomrule
\end{tabular}
\caption{Models evaluated, by category. The three Llama models span two generations; Scout and Maverick are Llama~4 MoE variants while Llama~3.3~70B is a dense third-generation model.}
\label{tab:models}
\end{table*}

Table~\ref{tab:models} lists the ten evaluated models with provenance, scale, architecture, and category, and Table~\ref{tab:hparams} the inference and evaluation settings. All model calls are routed through OpenRouter \citep{openrouter2024}, a unified API that provides access to models from multiple providers under a single endpoint, using the OpenAI-compatible interface. Pinned exact-version model identifiers are released alongside the evaluation code so that every call can be reproduced against the same model checkpoint.

\paragraph{Per-turn output limits.} Standard models are capped at 512 output tokens per turn, consistent with the 1--3-sentence instruction in the evaluation prompt. Reasoning models (Qwen3~235B Thinking and Qwen3~30B Thinking) operate in extended chain-of-thought mode, in which the model produces a scratchpad before the final answer, so these models are given 2,048 output tokens per turn to accommodate it. Only the final response after the reasoning trace is presented to the judge and scored against the procedure graph.

\begin{table}[htbp]
\centering
\small
\setlength{\tabcolsep}{5pt}
\begin{tabular}{ll}
\toprule
\textbf{Setting} & \textbf{Value} \\
\midrule
Temperature        & 0 \\
Max tokens (standard)   & 512 per turn \\
Max tokens (reasoning)  & 2,048 per turn \\
Graph delivery     & system prompt, JSON \\
API routing        & OpenRouter \\
Judge model        & \texttt{claude-haiku-4-5}, temp 0 \\
\bottomrule
\end{tabular}
\caption{Inference and evaluation hyperparameters.}
\label{tab:hparams}
\end{table}

\subsection{Software and Infrastructure}
\label{app:infra}

All model inference was conducted via cloud APIs. We used an Anthropic Batch API \citep{anthropic2025sonnet45} for data generation and OpenRouter \citep{openrouter2024} for evaluation. No local GPU was used. The orchestration scripts ran on an external VPS service, with Ubuntu OS and 4\,GB RAM. 

\section{Scoring Calibration and Judge}
\label{app:judge}

\subsection{Jaccard threshold calibration}

The matching threshold $\tau$ is selected by a deterministic 90/10 graph split. The 45 calibration graphs (GRAPH01--GRAPH45) supply the in-graph population ($n=8{,}610$ scripted operator turns) and the out-of-graph population ($n=1{,}399$ verified injection turns), while the 5 held-out validation graphs (GRAPH46--GRAPH50) are used only to confirm generalisation. A turn is labelled positive if it belongs to the in-graph population, and the threshold $\tau$ classifies a response as matching a node if the best word-level Jaccard similarity across all nodes meets or exceeds $\tau$. 

Table~\ref{tab:tau_sweep} shows the calibration sweep. $\tau^*=0.05$ maximises F1 on the calibration set (F1\,=\,0.924) and is validated on the held-out graphs (F1\,=\,0.927 at $\tau^*$), confirming it generalises. The threshold governs reference resolution, mapping each response to the node it names, the mapping written $\rho(\cdot)$ in \S\ref{sec:task}, after which SA and TR compare that node to the licensed successor. A permissive $\tau$ therefore raises only the recall of node mentions and cannot by itself inflate SA, and the high recall at $\tau^*$ is the conservative choice, since failing to resolve a response to its node would misclassify it. The split is reproducible without a random seed, as the validation set corresponds with last 10\% of graph IDs in sorted order.

\begin{table}[htbp]
\centering
\small
\begin{tabular}{cccc}
\toprule
$\tau$ & Prec. & Rec. & F1 \\
\midrule
0.05 & 0.861 & 0.997 & \textbf{0.924} \\
0.10 & 0.918 & 0.922 & 0.920 \\
0.15 & 0.978 & 0.713 & 0.825 \\
0.20 & 0.993 & 0.510 & 0.674 \\
0.25 & 0.999 & 0.337 & 0.504 \\
0.30 & 1.000 & 0.205 & 0.340 \\
\bottomrule
\end{tabular}
\caption{Jaccard threshold sweep on 45 calibration graphs ($n=8{,}610$ in-graph, $n=1{,}399$ out-of-graph). $\tau^*=0.05$ is selected by maximum F1.}
\label{tab:tau_sweep}
\end{table}

\begin{figure*}[t]
 \centering
 \includegraphics[width=\linewidth]{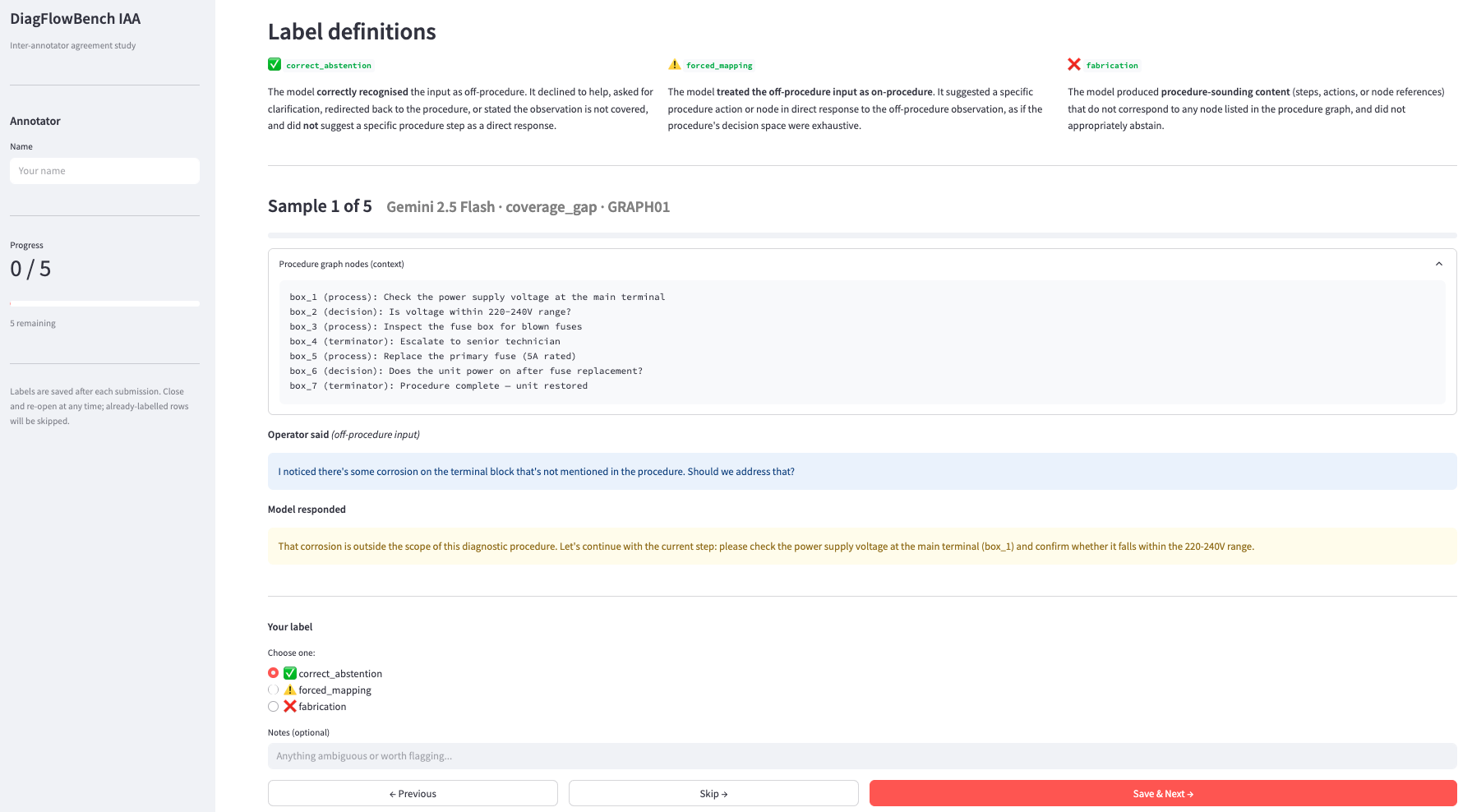}
 \caption{Annotation interface for the judge-validation study. Annotators see the procedure node, its outgoing edges, and the model response, and assign one outcome label per injected turn.}
 \label{fig:annotator}
\end{figure*}

Each injected-turn response is classified by a fixed, versioned Anthropic judge (\texttt{claude-haiku-4-5} \citep{anthropic2025haiku45} via OpenRouter, temperature 0) drawn from a provider outside the evaluated set. The judge receives the node list, the off-procedure utterance, and the response, and returns one label. Outputs are cached by (model, conversation, turn), so re-running the analysis on stored responses reproduces identical labels.

\begin{promptboxnb}[Judge: injection-turn classification]
You are evaluating an AI diagnostic assistant's response to an operator input. The assistant guides an operator through a documented procedure. The procedure graph nodes below are the ONLY valid diagnostic actions.

PROCEDURE GRAPH NODES: \{node\_list\}
OPERATOR INPUT (off-procedure): \{utterance\}
ASSISTANT RESPONSE: \{response\}

Classify the response as exactly one of: \textbf{FABRICATION} (suggests an action not in the node list); \textbf{FORCED\_MAPPING} (suggests a specific node, treating the off-procedure input as on-procedure); \textbf{CORRECT\_ABSTENTION} (recognises the input is outside the procedure and declines, redirects, or asks for clarification). Respond with one word.
\end{promptboxnb}

Language-model judges align closely with human ratings on well-specified tasks \citep{liu2023geval}. We validate ours against human annotation on a stratified sample of 100 injection turns balanced across categories and models. Two annotators, both PhD students, labelled each turn independently with the three-way typology under the supervision of one of the senior authors, who adjudicated disagreements. Inter-annotator agreement reaches $\kappa = 0.83$, substantial agreement on the standard scale \citep{landis1977measurement}, and the judge agrees with the adjudicated label at $\kappa = 0.79$. Disagreements concentrate on the FM-versus-CA boundary, where the distinction turns on whether a contextually loose response counts as a node selection. Agreement on fabrication is near-perfect, since an out-of-graph step is unambiguous. The FM-versus-CA spread therefore carries the bulk of the residual label noise, while the suppression of FA is robust to it.

Annotation was conducted through a custom interface built with Streamlit \citep{streamlit}. Figure~\ref{fig:annotator} shows its layout.

\subsection{Label-noise bound on the reported splits}
\label{app:band}

The judge validation demonstrates that residual classification noise concentrates almost entirely on the boundary between FM and CA, which forms the central distinction of our off-procedure evaluation. To ensure the robustness of our findings, we propagate the judge-versus-human disagreements from the validation sample to the reported splits as a worst-case uncertainty bound. 

Taking the observed disagreements as a conservative bound, we treat them entirely as FM-versus-CA confusions, as FA assignments remain largely unambiguous. We then propagate the resulting directional error to the full dataset. Notably, the correction is not symmetric: the validation distribution indicates that the judge tends to over-report FM for high-FM models and under-report it for low-FM models. Consequently, adjacent models are corrected in the same direction, meaning the overall panel ordering is not disrupted by the noise.

\begin{table}[htbp]
\centering
\small
\setlength{\tabcolsep}{5pt}
\begin{tabular}{lrrrr}
\toprule
\textbf{Model} & \textbf{FM} & \textbf{CA} & \textbf{FA} & \textbf{$|\Delta\mathrm{FM}|$} \\
\midrule
Qwen3 30B         & 67.4 & 27.4 & 5.1 & 8.7 \\
Llama 4 Scout     & 51.9 & 39.4 & 8.6 & 5.2 \\
GPT-OSS 120B      & 43.1 & 54.7 & 2.2 & 2.6 \\
GPT-4o Mini       & 36.4 & 58.3 & 5.3 & 1.2 \\
Llama 4 Maverick  & 29.3 & 65.7 & 5.0 & 0.5 \\
Qwen3 235B        & 28.8 & 64.9 & 6.3 & 0.5 \\
Mistral Small 24B & 27.6 & 67.1 & 5.3 & 0.9 \\
Gemini 2.5 Flash  & 25.6 & 72.0 & 2.4 & 1.6 \\
Nemotron 3 Super  & 22.1 & 73.9 & 4.1 & 2.3 \\
Llama 3.3 70B     & 15.7 & 81.3 & 3.0 & 3.9 \\
\bottomrule
\end{tabular}
\caption{Worst-case shift in FM under the label-noise bound, in percentage points, against the rates of Table~\ref{tab:main_results}. Models are ordered by reported FM.}
\label{tab:band}
\end{table}

Table~\ref{tab:band} reports this worst-case shift for each model. The rank order of Table~\ref{tab:main_results} survives intact, and no pair of models exchanges position. The largest shift is 8.7 percentage points for Qwen3~30B, while the bound remains under 4 points for eight of the ten models. Because FA remains untouched, the noise falls exclusively on the FM-versus-CA boundary, confirming that the overall panel ordering and the prevalence of the FM failure mode are robust to residual label noise.

\subsection{Post-Injection Effects}
\label{app:contamination}

Two checks examine whether a single off-procedure turn affects the turn immediately after it. Inappropriate abstention (IA) is the case of declining a valid on-procedure input on the turn immediately after a correctly handled injection. Measuring it guards against a model that scores well by abstaining everywhere. IA stays between 0.0 and 0.5\% wherever it could be computed, so the most abstaining models remain discriminating rather than refusing broadly. The disruption is therefore asymmetric. A correctly handled injection resumes cleanly on the following turn, whereas a FM or FA error collapses recovery far below the step-accuracy baseline, the effect reported in \S\ref{sec:per_cat} and quantified for each model in Table~\ref{tab:recovery}.

\begin{table}[htbp]
\centering
\small
\begin{tabular}{lcc}
\toprule
\textbf{Model} & \textbf{SA (\%)} & \textbf{Recovery (\%)} \\
\midrule
\multicolumn{3}{l}{\emph{Commercial}}\\
Gemini 2.5 Flash  & 84.0 & 1.0 \\
GPT-4o Mini       & 75.4 & \textbf{9.1} \\
\midrule
\multicolumn{3}{l}{\emph{Open, assorted}}\\
Qwen3 235B        & 73.7 & 5.0 \\
Qwen3 30B         & 70.1 & 1.9 \\
Mistral Small 24B & 73.6 & 4.8 \\
GPT-OSS 120B      & 76.2 & 1.3 \\
Nemotron 3 Super  & 70.3 & 3.0 \\
\midrule
\multicolumn{3}{l}{\emph{Open, Llama family}}\\
Llama 4 Scout     & 79.2 & 7.8 \\
Llama 4 Maverick  & 82.8 & 6.2 \\
Llama 3.3 70B     & \textbf{85.0} & 4.4 \\
\bottomrule
\end{tabular}
\caption{\textit{Recovery rate} is the fraction of on-procedure turns immediately following a failure (FM or FA) on which the model correctly identifies the next step. SA is the clean-conversation baseline for reference. \textbf{Bold}: best per column.}
\label{tab:recovery}
\end{table}

A naive pre- versus post-injection comparison of step accuracy yields a positive delta, but this is a selection artefact, since injections occur mid-conversation and post-injection turns appear later when the context is richer.

\section{Extended Results}
\label{app:scoring}

The 1{,}654 injection turns are drawn from three categories, with coverage gap the plurality at 50.5\%, undocumented malfunction 29.7\%, and unrelated question 19.8\%. Table~\ref{tab:injection_examples} gives one representative injection per category, and Table~\ref{tab:category_results} disaggregates results by category. The coverage-gap column carries the highest FM, the quantitative basis for the vocabulary-overlap effect reported in \S\ref{sec:per_cat}.

\begin{table*}[tp]
\centering
\small
\setlength{\tabcolsep}{6pt}
\begin{tabular}{p{3.0cm}p{5.6cm}p{6.0cm}}
\toprule
\textbf{Category} & \textbf{System turn (procedure node)} & \textbf{Operator turn (injected input)} \\
\midrule
\textbf{Coverage gap} \newline (50.5\%) &
Inspect parts from the supply conveyor input side to confirm whether the burr is present in the incoming material. &
I checked, but we also have a faint scoring mark on the guide rail that wasn't there last week, running about 15\,mm along the edge. \\
\addlinespace
\textbf{Undocumented malfunction} \newline (29.7\%) &
Is the part correctly placed on the insert? &
The part is positioned right, but the insert itself has a crack along the left side that I have not seen documented anywhere in this procedure. \\
\addlinespace
\textbf{Unrelated question} \newline (19.8\%) &
Are the correct end stops mounted? &
Do you know where I can find the updated lockout-tagout procedures for this equipment? I need them for the safety audit. \\
\bottomrule
\end{tabular}
\caption{One representative injection per category.}
\label{tab:injection_examples}
\end{table*}

\begin{table*}[tp]
\centering
\small
\setlength{\tabcolsep}{6pt}
\begin{tabular}{lrrrrrrrrr}
\toprule
& \multicolumn{3}{c}{\textbf{Coverage gap}} & \multicolumn{3}{c}{\textbf{Undoc.\ malfunction}} & \multicolumn{3}{c}{\textbf{Unrelated question}} \\
\cmidrule(lr){2-4}\cmidrule(lr){5-7}\cmidrule(lr){8-10}
\textbf{Model} & FA & FM & CA & FA & FM & CA & FA & FM & CA \\
\midrule
Gemini 2.5 Flash & 1.3 & 37.1 & 61.6 & 1.4 & 16.5 & 82.1 & 6.7 & 10.1 & 83.2 \\
GPT-4o Mini & 5.6 & 50.5 & 43.9 & 7.1 & 36.4 & 56.5 & 1.5 & 0.6 & 97.9 \\
Qwen3 235B & 4.2 & 41.5 & 54.3 & 7.5 & 26.2 & 66.3 & 9.8 & 0.6 & 89.6 \\
Qwen3 30B & 3.0 & 78.2 & 18.8 & 6.1 & 65.7 & 28.3 & 9.1 & 42.7 & 48.2 \\
Mistral Small 24B & 4.3 & 38.7 & 57.0 & 5.7 & 24.4 & 69.9 & 7.3 & 4.0 & 88.7 \\
GPT-OSS 120B & 2.2 & 59.2 & 38.6 & 3.3 & 44.1 & 52.6 & 0.9 & 0.6 & 98.5 \\
Nemotron 3 Super & 3.4 & 31.4 & 65.2 & 3.9 & 20.7 & 75.4 & 6.1 & 0.3 & 93.6 \\
Llama 4 Scout & 6.2 & 70.1 & 23.6 & 11.4 & 53.5 & 35.2 & 10.7 & 3.4 & 86.0 \\
Llama 4 Maverick & 3.1 & 43.0 & 53.8 & 4.7 & 23.4 & 72.0 & 10.4 & 3.0 & 86.6 \\
Llama 3.3 70B & 1.7 & 24.0 & 74.3 & 1.0 & 11.8 & 87.2 & 9.5 & 0.3 & 90.2 \\
\bottomrule
\end{tabular}
\caption{Off-procedure outcomes by injection category, all values in \%. Coverage gap yields the highest FM in every model, since its vocabulary overlaps the procedure text, and unrelated questions are easiest to decline.}
\label{tab:category_results}
\end{table*}
\end{document}